\documentclass{llncs}
\usepackage{caption}
\usepackage{algorithm, float, graphicx, mathtools, multirow, array, makecell, hhline, tabularx, amssymb, amsmath, subcaption}

\usepackage{xcolor}

\usepackage{colortbl}
\usepackage{graphicx}
\usepackage{hyperref}
\usepackage{booktabs}
\usepackage{siunitx}
\usepackage{etoolbox}
\usepackage{enumitem}
\hypersetup{urlcolor=blue}
\usepackage{amsmath}
\usepackage{dsfont}
\usepackage{tikz,soul}
\hypersetup{colorlinks=true,urlcolor=blue}
\usepackage[noend]{algpseudocode}%
\usepackage[para]{footmisc}
\setlist[itemize]{topsep=0pt, partopsep=0pt, itemsep=0pt, parsep=0pt}
\setlist[itemize,enumerate]{topsep=0pt, partopsep=0pt, itemsep=0pt, parsep=0pt}
\newcommand{\cellcolorbasedonvalue}[2]{%
    \cellcolor{green!#1!white}%
    \ifnum#1>50 \color{black} #2 
    \else \color{black} #2 
    \fi
}

\usepackage{algorithm, float, graphicx, mathtools, multirow, array, makecell, hhline, tabularx, amssymb, amsmath, subcaption}

\usepackage{xcolor}
\usepackage{graphicx}
\usepackage{hyperref}
\usepackage{booktabs}
\usepackage{siunitx}
\usepackage{etoolbox}
\hypersetup{urlcolor=blue}
\usepackage{amsmath}
\usepackage{dsfont}
\hypersetup{colorlinks=true,urlcolor=blue}
\usepackage[noend]{algpseudocode}%
\usepackage{orcidlink}
\newcolumntype{C}[1]{>{\centering\let\newline\\\arraybackslash}p{#1}}%
\newcolumntype{L}[1]{>{\raggedleft\let\newline\\\arraybackslash}p{#1}}%
\newcolumntype{R}[1]{>{\raggedright\let\newline\\\arraybackslash}p{#1}}%
\newcommand{\cma}{{\small\textcolor{blue}{CMA}}}
\newcommand{\de}{{\small\textcolor{blue}{DE}}}
\newcommand{\rs}{{\small\textcolor{blue}{RS}}}
\newcommand{\mbh}{{\small\textcolor{blue}{MBH}}}

\begin{document}%
\title{Stalling in Space: Attractor Analysis for any Algorithm}

\author{Sarah L. Thomson$^{1}$\orcidlink{0000-0001-6971-7817} \and Quentin Renau$^{1}$\orcidlink{0000-0002-2487-981X} \and Diederick Vermetten$^{2}$\orcidlink{0000-0003-3040-7162}\and Emma Hart$^{1}$\orcidlink{0000-0002-5405-4413} \and Niki van Stein$^{2}$\orcidlink{0000-0002-0013-7969} \and Anna V. Kononova$^{2}$\orcidlink{0000-0002-4138-7024}
\institute{\inst{1}Edinburgh Napier University, UK
\email{s.thomson4@napier.ac.uk}
\institute{2} LIACS, Leiden University, The Netherlands}
}

\maketitle
\begin{abstract}
\sloppypar
Network-based representations of fitness landscapes have grown in popularity in the past decade; this is probably because of growing interest in \emph{explainability} for optimisation algorithms. Local optima networks (LONs) have been especially dominant in the literature and capture an approximation of local optima and their connectivity in the landscape. However, thus far, LONs have been constructed according to a strict definition of what a local optimum is: the result of local search. Many evolutionary approaches do not include this, however. Popular algorithms such as CMA-ES have therefore never been subject to LON analysis. Search trajectory networks (STNs) offer a possible alternative: nodes can be any search space location. However, STNs are not typically modelled in such a way that models temporal \emph{stalls}: that is, a region in the search space where an algorithm fails to find a better solution over a defined period of time. In this work, we approach this by systematically analysing a special case of STN which we name \emph{attractor networks}. These offer a coarse-grained view of algorithm behaviour with a singular focus on stall locations. We construct attractor networks for CMA-ES, differential evolution, and random search for 24 noiseless black-box optimisation benchmark problems. The properties of attractor networks are systematically explored. They are also visualised and compared to traditional LONs and STN models. We find that attractor networks facilitate insights into algorithm behaviour which other models cannot, and we advocate for the consideration of attractor analysis even for algorithms which do not include local search. 
\end{abstract}%
\section{Introduction}%

Understanding the landscape or structure of the search space associated with a problem instance is important for both predicting the performance of algorithms and improving algorithm design so that they can more efficiently traverse the landscape.
In the past decade, several techniques for visualising landscapes have arisen that provide new insights into this task.  Local optima networks (LONs)~\cite{ochoa2008study} capture the number, distribution and connectivity pattern of local optima in the form of a network where nodes represent local optima and edges capture the transitions between them. LONs have most often been used to understand landscapes in combinatorial settings~\cite{treimun2020modelling,ochoa2017understanding,ochoa2019local} --- mainly due to the fact that constructing a LON requires an iterated local search (ILS) algorithm to be run to identify local optima. More recently, LONs have been constructed for continuous spaces, mostly using monotonic basin-hopping~\cite{adair2019local,mitchell2023local}, which is essentially an ILS framework for continuous optimisation. The LON model has been particularly useful identifying basins of attraction in a landscape~\cite{ochoa2019local,sanchez2024regularized}. Search trajectory networks (STNs)~\cite{ochoa2020search} were introduced as a more generalisable alternative to LONs. STN are directly constructed from data gathered while running an algorithm. Unlike LONs, there is no condition that nodes must be a local optima. STNs enable a user to visualise (for example) whether multiple algorithms or runs of an algorithm traverse the same locations, indicate termination points of an algorithm (both optimal and sub-optimal) and show the frequency with which algorithms pass through and escape from the nodes via an edge weight. However, STNs do not typically provide information as to \textbf{how long} an algorithm remained stalled at a node before managing to escape --- the information they encode is valuable, but does not generally have this temporal aspect. 

To address this, we put forward a new variant of an STN which is applicable to any type of algorithm (including those that do not easily permit local search) and which demonstrate the \textit{stagnation behaviour} of an algorithm at various points in the search. We dub these regions \emph{stall locations}: a period of the search process where the best value of the objective function does not change over a period of $\beta$ evaluations. A stall location can hence be seen as an attractor of the search and can be detected using any type of algorithm (i.e. it does not rely on local search). This approach, which we call \textit{attractor networks}, brings new insights into the operation of algorithms which are not evident in either LON networks or STN networks. We study the properties and illustrate the benefits of attractor networks by constructing them using three different algorithms on the 24 functions of the BBOB test suite~\cite{cocoJournal} --- comparing metrics and visualisations obtained from the networks to those obtained from both LON and STN. The contributions can be summarised as follows:
\begin{itemize}
    \item The formal introduction of a method for visualising and analysing attractors in the genotype space (termed an \textit{attractor network, or AN}) in a landscape that is applicable to algorithms from both the continuous and combinatorial optimisation domains and regardless of whether local search is applicable.
    \item A demonstration that ANs can provide new insights into intermediate attractors in the search process that are not detected by standard LON or STN analysis, obtained by evidence gathered by conducting a systematic analysis over suite of functions and algorithms.
\end{itemize}
\section{Background}%
\subsection{Network models}

\paragraph{Local optima network (LON).} A local optima network is a directed and weighted graph comprised of: a) nodes $lo_i \in LO$ which belong to a set of local optima (this can be the complete set or a sample) and b) edges $e_{ij} \in E$ which connect pairs of local optima (nodes) $lo_i$ and $lo_j$ with a weight which denotes the frequency of transition $w_{ij}$ iff $w_{ij} > 0$. Extensive descriptions can be found in~\cite{verel2012local}.

\paragraph{Search trajectory network (STN).} A search trajectory network is a directed and weighted graph comprised of: a) nodes $sl_i \in SL$ which belong to a set of search locations from algorithm trajectories (these do not need to be local optima) and b) edges $e_{ij} \in E$ which connect pairs of locations (nodes) $sl_i$ and $sl_j$ with a weight which denotes the frequency of transition $w_{ij}$ iff $w_{ij} > 0$. Extensive descriptions can be found in~\cite{ochoa2021search}.

\subsection{LON and STN manifestations}

In combinatorial spaces, LONs are constructed such that a local optimum is the result of the application of a local search method~\cite{verel2012local}. In this way, the LON nodes satisfy the condition that they are the best solution within their [sampled or complete] neighbourhood with respect to a basic mutation operator. This is usually achieved either by using iterated local search as the construction algorithm~\cite{treimun2020modelling,ochoa2017understanding,ochoa2019local} or occasionally by augmenting population-based approaches with local search, rendering them memetic~\cite{veerapen2016tunnelling,thomson2020local}. 

Approaches in continuous spaces are typically similar; the Limited-memory Boyden-Goldfarb-Shanno (L-BFGS-B)~\cite{wright2006numerical} optimiser --- a quasi-Newton method --- has been used as local search to obtain LON nodes~\cite{adair2019local,mitchell2023local,contreras2020synthetic}. The Nelder-Mead
downhill simplex algorithm has been used as an alternative method to identify local optima for LONs~\cite{karatas2021towards}; very recently, a greedy local sampling akin to \((1 + \lambda)\) Evolutionary Strategy has been proposed for discovering local optima for these purposes~\cite{fieldsend2024scalable}. One contribution augmented differential evolution with local search so that LONs could be constructed~\cite{Homolya2019}. The common thread with LON works in continuous spaces is that the network is constructed using algorithms which may not always closely resemble the type of algorithms actually used to search these kind of spaces in practice. For example, networks which reflect the dynamics encountered by monotonic basin-hopping may only give \textit{limited insight} into how CMA-ES behaves on a problem. 

STNs are more generalisable than LONs, but have their own considerations. One is the decision of when to log a new STN node. As it relates to population-based algorithms moving in continuous spaces, the convention is logging the best individual in the population every $G$ generations; $G$ has been set as (for example) 1~\cite{ochoa2021search} or the dimension of the problem~\cite{ochoa2020search}. This could potentially have the effect that a location is logged even if the search is only there for as little as one generation. 

Very recently, an article on Cartesian genetic programming (CGP)~\cite{de2024multimodal} took the alternative approach of logging STN nodes if the search was at a location in the behaviour space for at least one iteration. However, in practice, the search was usually stuck at locations in the behaviour space for many iterations. It is therefore the case that what resulted was essentially an attractor network (as it is defined in this paper), although there are some \textit{key differences}: a) this was not in the genotype space, as we consider here; b) a single node in the behaviour network represented typically thousands of genotypically different solutions from the genotype space which map to the same program behaviour; and c) the condition for a behaviour location to be considered a stall point was a single iteration. Despite the differences with the networks we analyse here, in that work attractors were visualised and this helped give valuable comparative insights about the two studied CGP algorithms. This finding serves as a good motivator for our work in focussing on attractors.
 
Another choice related to STN modelling in continuous spaces is what the threshold in decision space should be defined to consider two solutions the same. For this, a \textit{partition factor} has been used~\cite{ochoa2020search,ochoa2021search}. The same concept has also been used in LON analysis for continuous functions; the original paper used a threshold of \(10^{-5}\)~\cite{adair2019local} and this is also the value used in the Python package \textsc{pflacco}~\cite{prager2024pflacco}. With STNs, the partition factor has conventionally been decided according to a formula involving the search space bounds, $x_{max}$ and $x_{min}$ and dimension of the problem. They consider the largest integer $n$ for which the following holds: \((x_{\text{max}} - x_{\text{min}}) \times D \geq 10^n\); with $n$, the expression \(2 - n\) is then carried out to obtain the partition factor, which is equivalent to the exponent of 10. According to this formula, our partition factors would lead to coordinate precision $\epsilon$ = \(10^{-1}\) for 2D and $\epsilon$ = \(10^{0}\) for 10D. This is rather coarse when we consider that our region of interest is [-5, 5]; we therefore take \(10^{-2}\) as our most coarse setting, but also consider other lower values which are specified in Section~\ref{sec:setup}.

It is standard with visualisation of STNs to make the size of the nodes proportional to how many search runs reached that location~\cite{chacon2023search} and the related notion of \emph{attraction areas} has been considered recently~\cite{chacon2024large}; however, these ideas are \textit{not} equivalent to the notion of attractor which we use in this paper. The visual node sizes and identified attraction areas will show locations where search goes frequently, \textbf{but not how long it stays there}. For example: a node which is large in a standard STN visualisation could represent a location where search flows through often, but where it does not stay for longer than a single generation of the algorithm. There could also be a case where a node is small --- due to having only been found in a single run --- but search stalled there for a significant number of evaluations before finally moving on. This would not be captured with standard STN conventions, but is important to capture. In general, neither LONs nor STNs typically consider the temporal aspect of algorithm behaviour directly. That thought, alongside an appetite for applying LON analysis to real-world problems (where the relevant search algorithms may not mirror LON construction methods, but attractors may nevertheless be of great importance), motivates the idea of attractor networks introduced here. 

\subsection{The Notion of Attractors for CMA-ES and DE} 
As introduced in Section~\ref{sec:setup}, experiments carried out in this study are based on variants of two popular heuristic algorithms for continuous setting: CMA-ES and DE. Therefore, we briefly examine the literature for the analysis of attractors and these two algorithm classes.

To the best of our efforts, no papers were identified that consider search behaviour in the sense of stalling moments or locations during runs of continuous optimizers. While not really discussing the concept of stalling/attractors per se, the heuristic (continuous) optimisation community has actively worked on mechanisms preventing such stalling. These include: restarts of stagnated~\cite{ipop} or redundant~\cite{deNobel2024PPSN} runs with possible increase of population size (IPOP~\cite{ipop} and BIPOP~\cite{bipop} mechanisms in CMA-ES and adaptive population control~\cite{Tanabe2013SHADE,Tanabe2014LSHADE} in DE), step-size thresholding~\cite{Hansen2001} and control of covariance matrix eigenvalues in CMA-ES, various population diversity improving components for both algorithms, improved covariance adaptation mechanisms for high-conditioned landscapes, improved handling of boundary constraints to balance exploration of the whole domain~\cite{Kononova2024tiobr}. 

\section{Methodology}\label{methods}

\subsection{Attractor Networks}
In the context of optimisation algorithms, the notion of attractors can be defined in a traditional sense inspired by~\cite{milnor1985concept,collet2009iterated} that is independent of the algorithm, as described in~\cite{AntonovSPIE2023}. For a metric space \( X \), the basin of attraction in discrete-time dynamics is defined for the system \( (X, \varphi: X \to X) \), where the \( k \)-th iteration of the system corresponds to the \( k \)-fold composition \( \varphi \circ \varphi \circ \dotsc \circ \varphi (\cdot) \). The application of \( \varphi \) to \( x \) is assumed to shift \( x \) towards a vector of smaller magnitude in the direction opposite to \( \nabla f(x) \). If this iterative process converges to a single point \( x \) such that \( \varphi(x) = x \), this point is referred to as an \textit{attractor}.

\paragraph{Attractor.} In the context of attractor networks, we define an \emph{attractor} to be a location in the search space where at least one of the $r$ algorithm runs used to construct the network stalled for at least $\beta$ fitness evaluations. We are working in continuous spaces in this study; it follows that the notion of attractor also depends on the coordinate precision $\epsilon$. This value mandates the cutoff threshold in decision space for two solutions to be considered the same attractor node in the network: if the two solutions differ in at least one variable by an amount equal to or over $\epsilon$ then they will not be represented by the same node.  

\paragraph{Attractor network construction.} Attractor networks can be constructed from a log taken during algorithm runs. The log must include all moments of improvement to the best-so-far solution; each logging event will consist of three things: the number of elapsed fitness evaluations, the fitness of the new solution, and the genotype of that solution. From this type of log, an attractor network can be built in mostly the same manner as an STN, except nodes (and edges) are \textbf{exclusively} logged if the number of elapsed evaluations between two consecutive improvements to the best-so-far solution is greater than $\beta$. Mirroring LONs and STNs, an attractor network is comprised of the trajectories for multiple independent runs of the algorithm. The network is also defined according to a level of coordinate precision $\epsilon$: the radius in decision space [computed with Manhattan distance] within which two solutions are considered the same node. 

\paragraph{A note on attractor networks.}

We note that standard LONs are a special case of STNs, essentially encoding search trajectories of monotonic basin-hopping (in continuous space) or iterated local search (combinatorial). The attractor networks described here may be seen as both LONs and STNs --- although, instead of strict hill climbing local optima we consider generalised attractors. The perspective is non-specific enough that any algorithm can be studied using it, unlike LONs. STNs can also be used to understand any algorithm, but they have not been modelled in such a way that attractor structure in the genotype space is evident. Attractor networks could be viewed as a reduced STN, and essentially offer a coarser-grained view with a singular focus on search traps.

\subsection{Approach to validation}

Naturally, the attractor networks should be compared with standard local optima networks and standard search trajectory networks. This brings to mind the question of what {\lq{standard}\rq} means for these models; we now clearly define what we consider to be the standard models (for the purposes of the study). We define the standard LON as the method implemented in \textsc{pflacco} and described fully in previous literature~\cite{adair2019local}. The construction is based on repeated runs of monotonic basin-hopping with L-BFGS as the local search component. We consider the standard STN to be such that a node (location) is logged as the best individual in the population every $k$ iterations, which is the setting used in a previous STN study on population-based algorithms for continuous domains~\cite{ochoa2020search}. For random search, this is every $k$ evaluations; for the other algorithms, it is every \(k 
\times popsize\) evaluations. 

In addition to comparison with other network models, random search is included in our portfolio for validation. Random search should stall frequently and show approximately the same behaviour across all functions. That is, random search attractor networks should not look significantly different depending on the function. In addition, the attractor networks of random search should not resemble those of our other algorithms. 

\section{Experimental Setup}\label{sec:setup}%

For our experiments, we use COCO's~\cite{coco_hansen2021} BBOB problem suite~\cite{bbob_hansen2009_noiseless}, consisting of 24 single-objective, noiseless, continuous minimisation problems, which we access via \textsc{IOHexperimenter}~\cite{iohexp}. We make use of a single instance (IID 1) for each problem, in both problem dimensionality 2 and 10. 
We consider several settings for $\beta$: $\in [10,$ $20,$ $40,$ $80,$ $160,$ $320,$ $640]$ and $\epsilon$ for the attractor networks: $p \in [0.01,$ $0.001,$ $0.0001,$ $0.00001]$. We construct attractor networks from 10 runs and 30 runs, respectively: the smaller networks are for visualisation, while the larger ones are used for computing statistics. For both the standard LON and standard STN, we consider the same $\epsilon$ for decision variables as the setting found in the standard LON implementation in the \textsc{pflacco}\footnote{version 1.2.2} package: \(10^{-5}\). LON construction is implemented through \textsc{pflacco} functions and the STN construction was written from scratch in Python (the code and data will be published upon acceptance of this work). LON extraction has a parameter which serves as the termination condition for an individual run: we use the default setting for this, which is 1000 iterations without improvement. To match the attractor networks, we also construct LONs and STNs from 10 and 30 runs, for visualisation and statistics purposes, respectively. In the cases where networks relating to 10-dimensional functions are plotted, the position of nodes on the \emph{x}-axis is obtained through \textsc{scikit-learn}~\cite{scikit-learn} multi-dimensional scaling on the decision vectors. Where search on 2-dimensional functions is visualised, the position of nodes is relative to their actual location in decision space. 

Our algorithm portfolio consists of three algorithms: vanilla CMA-ES from the ModCMA package~\cite{modcma}, DE rand/1/bin with uniform initialisation, $F=0.5$, $Cr=0.5$ and saturation corrections for infeasible solutions from ModDE~\cite{modde} and RandomSearch taken from Nevergrad~\cite{nevergrad}\footnote{Versions used: modcma 1.0.2 (C++ backend), modde 0.0.1, nevergrad 1.0.0.}. For both modular algorithm packages, we stick with the \textit{default parameter settings}, as we aim to illustrate the attractor network methodology rather than explore the best-performing versions of these algorithms. In particular, this decision means that the population sizes are dimensionality-dependent, with population sizes 6 and 10 for the 2 and 10-dimensional problems, respectively (which is relatively low for DE, but ensures having an equal amount of generations for both algorithms). From now on, we will refer to CMA-ES as {\cma}, DE as {\de}  and random search as {\rs}. In addition, the monotonic basin-hopping algorithm used in LON construction is referred to as {\mbh}.

\begin{figure*}[tb!]
        \centering
        \begin{subfigure}[b]{0.32\textwidth}
            \centering
            \includegraphics[trim=25 30 25 25,clip,width=\textwidth]{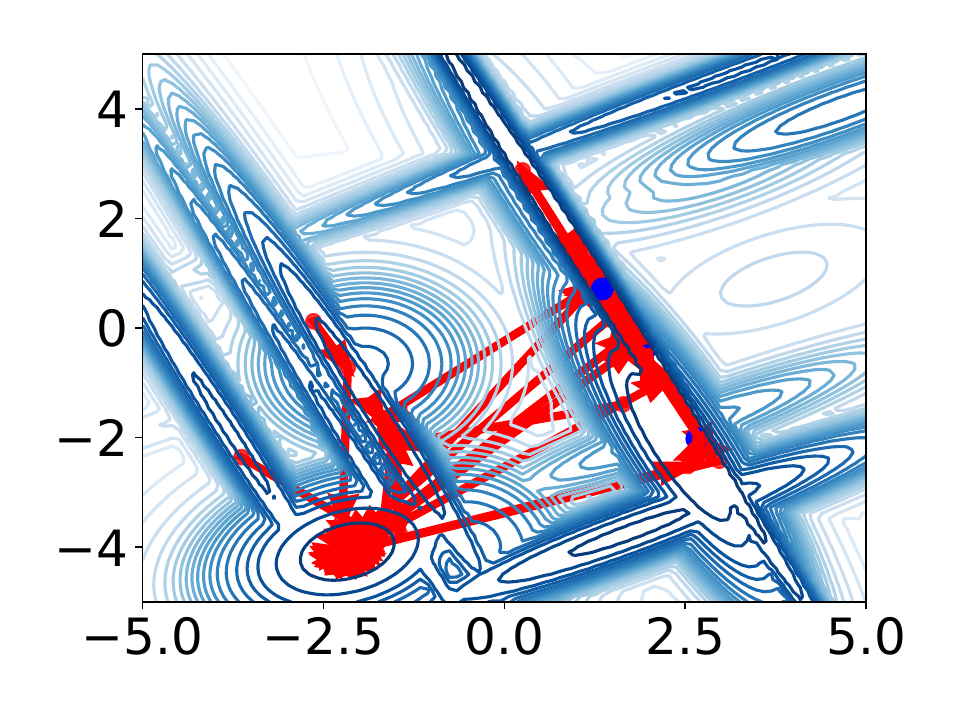}
            \caption{\(\beta\) = 10}
            \label{fig:f21-low}
        \end{subfigure}
        \begin{subfigure}[b]{0.32\textwidth}
            \centering
            \includegraphics[trim=25 30 25 25,clip,width=\textwidth]{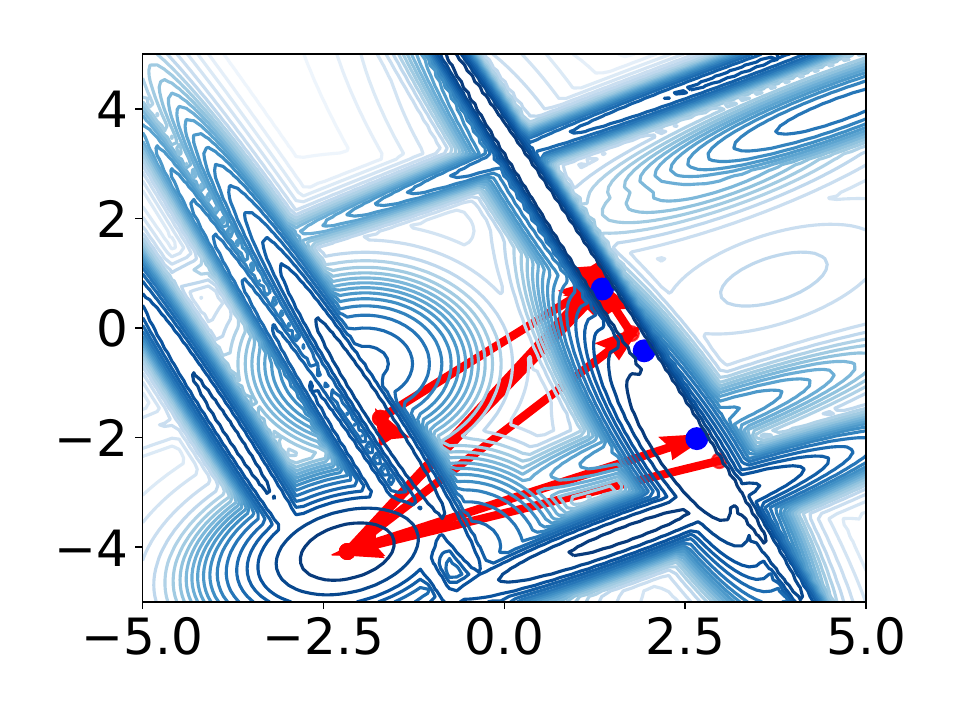}
            \caption{\(\beta\) = 80}
            \label{fig:f21-medium}
        \end{subfigure}
        \begin{subfigure}[b]{0.32\textwidth}  
            \centering 
            \includegraphics[trim=25 30 25 25,clip,width=\textwidth]{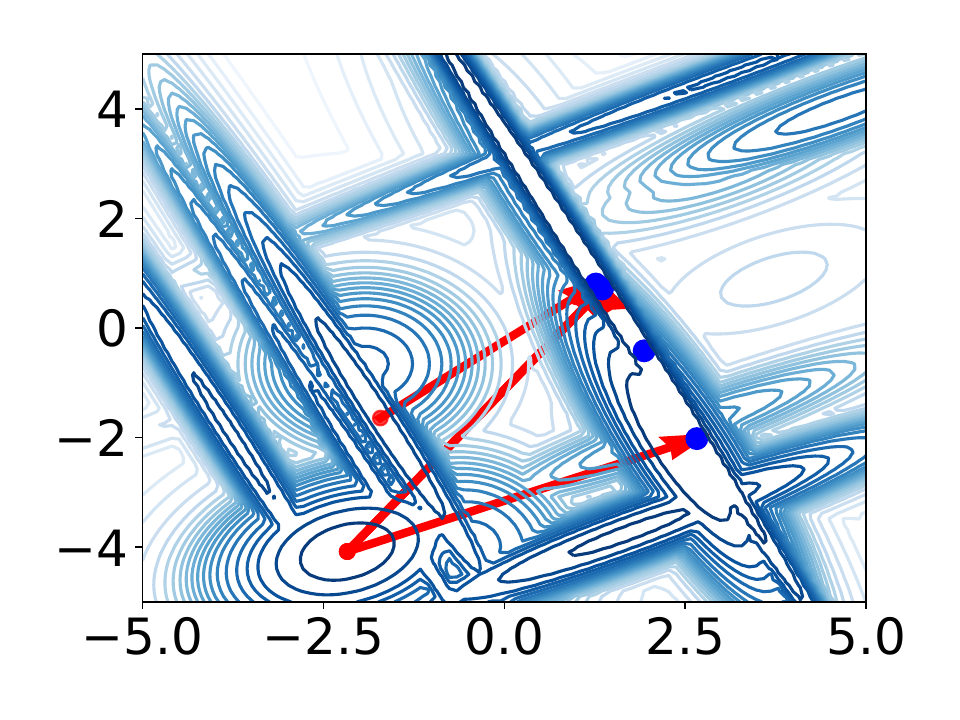}
            \caption{\(\beta\) = 320}   
            \label{fig:f21-high}
        \end{subfigure}
        \caption{Attractor networks for {\cma} on 2D Gallagher function \emph{f21} across different values of \(\beta\).}
            \label{fig:an-viz-2d-function-overlay}
    \end{figure*}%

\begin{figure*}[t!]
    \centering

        \begin{subfigure}[b]{0.3\textwidth}
            \centering
            \includegraphics[trim=70 35 30 50,clip,width=\textwidth]{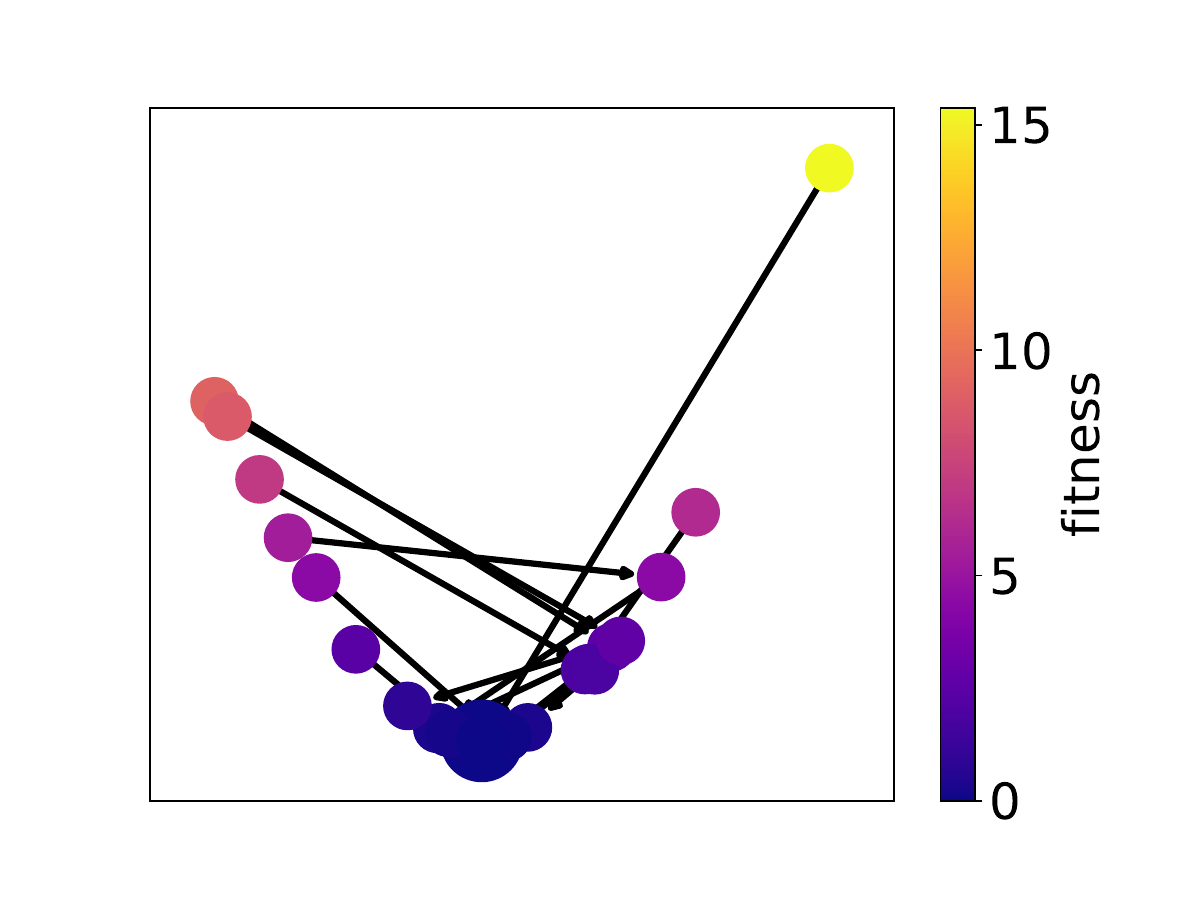}
            \caption{\(\beta = 40\)}
            \label{fig:beta40}
        \end{subfigure}
      \begin{subfigure}[b]{0.3\textwidth}
            \centering
           \includegraphics[trim=70 35 30 50,clip,width=\textwidth]{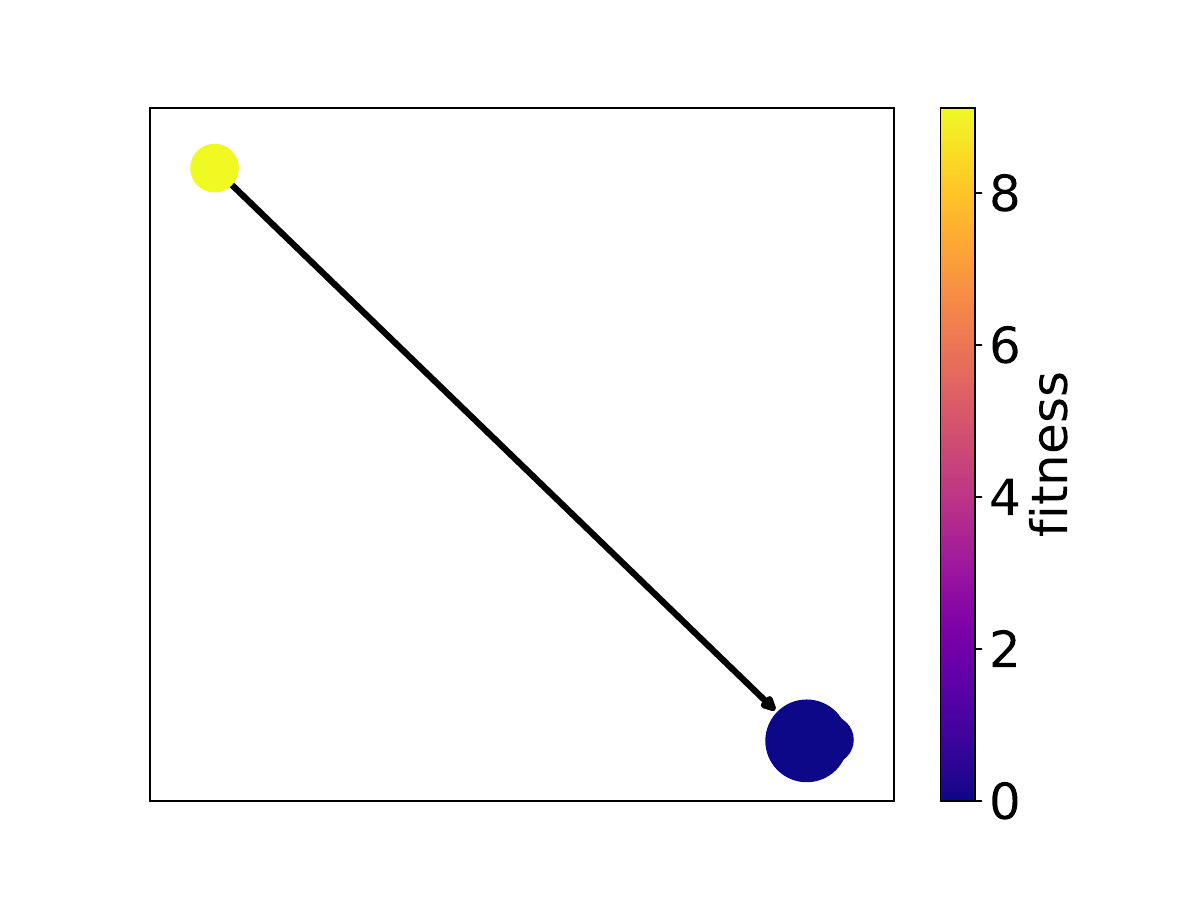}
            \caption{\(\beta = 80\)}
            \label{fig:beta80}
        \end{subfigure}
            \centering
             \begin{subfigure}[b]{0.3\textwidth}
           \includegraphics[trim=70 35 30 50,clip,width=\textwidth]{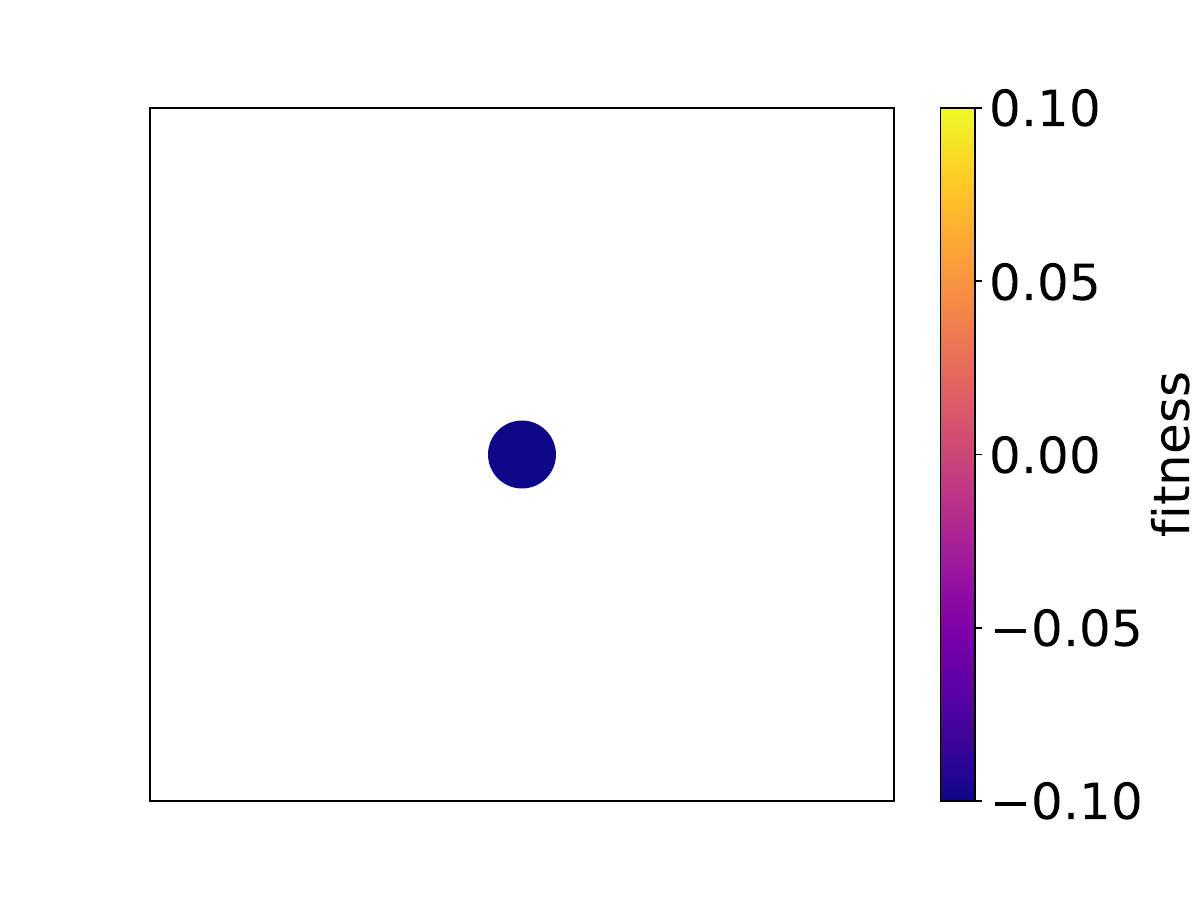}
            \caption{\(\beta = 160\)}
            \label{fig:beta160}
        \end{subfigure}       

    \caption{Attractor networks [$\epsilon$=\(10^{-5}\)] for {\cma} on the 10-dimensional Sphere function $f1$ across different values of \(\beta\). Node size is proportional to the number of runs which reached that location}
    \label{fig:sphere-beta}
\end{figure*}

\section{Results}%

\subsection{Attractor Networks: Characteristics}

We begin by visualising and analysing attractor networks. Figures~\ref{fig:an-viz-2d-function-overlay} and~\ref{fig:sphere-beta} show the change in {\cma} AN structure with increasing $\beta$ (left-to-right) for \emph{f21} and \emph{f1}, respectively. Figure~\ref{fig:an-viz-2d-function-overlay} reflects 2D function optimisation. The axes of the plots, and the placement of AN nodes, reflect the actual 2D coordinates; colour represents fitness. Figure~\ref{fig:sphere-beta} reflects search on 10D functions. In both Figures, we can see that increasing $\beta$ leads to sparser attractor networks. It is interesting that for both, there are still attractors even at a high $\beta$ --- in the case of Figure~\ref{fig:f21-high}, which represents optimisation with a population size of 6, we can see that there are still multiple {\cma} attractors even with $\beta$=320 fitness evaluations (which equates to 53 generations). Similarly, Figure~\ref{fig:beta80} shows that 10-dimensional \emph{f1} --- the uni-modal sphere function --- is associated with a sub-optimal attractor when $\beta$=80, which equates to 8 generations of {\cma}. We generated Figures for all combinations of [function, algorithm, dimension, $\beta$, and $\epsilon$]: these can be found in the supplemental material \cite{https://zenodo.org/records/14170241}.

\begin{table}[b!]
     \caption{Median and IQR value [over 24 functions] for the median evaluations differential [over all edges in a given AN with $\epsilon$=\(10^{-5}\) built from 30 runs]. Cells are shaded according to how often [out of 24 functions] the global optimum is present in the network: more vibrant green is more often}
      \label{tab:network-metrics}
      \centering
        \resizebox{0.85\textwidth}{!}{%
      \begin{tabular}{lcccc}
        \hline
        \multirow{2}{*}{\phantom{aaa}} & \multicolumn{2}{c}{2D} & \multicolumn{2}{c}{10D} \\
        \cline{2-5}
       model & median & IQR value & median & IQR value \\
         \hline
       {\de} AN [$\beta$=40] & \cellcolorbasedonvalue{33}{52.25} & \cellcolorbasedonvalue{33}{6.13} & \cellcolorbasedonvalue{4}{55.5} & \cellcolorbasedonvalue{4}{67.88} \\
       {\de} AN [$\beta$=80]  & \cellcolorbasedonvalue{33}{101}  & \cellcolorbasedonvalue{33}{15.5}  & \cellcolorbasedonvalue{4}{119.25} & \cellcolorbasedonvalue{4}{93.13}  \\  
       
       {\cma} AN [$\beta$=40] & \cellcolorbasedonvalue{96}{61.5}  & \cellcolorbasedonvalue{96}{13.25}  & \cellcolorbasedonvalue{50}{61.5} & \cellcolorbasedonvalue{50}{7.5}  \\
       {\cma} AN [$\beta$=80] & \cellcolorbasedonvalue{96}{168}  & \cellcolorbasedonvalue{96}{153.75}  & \cellcolorbasedonvalue{50}{108} & \cellcolorbasedonvalue{50}{18.75} \\

       {\rs} AN [$\beta$=40] & \cellcolorbasedonvalue{16}{509} & \cellcolorbasedonvalue{16}{128.63} & \cellcolorbasedonvalue{0}{484} & \cellcolorbasedonvalue{0}{126} \\
       {\rs} AN [$\beta$=80] & \cellcolorbasedonvalue{16}{732.25} & \cellcolorbasedonvalue{16}{132.88} & \cellcolorbasedonvalue{0}{659.5} & \cellcolorbasedonvalue{0}{191.5} \\
       \hline
    \end{tabular}}\label{tab:edgediff-metrics}
\end{table}

Table~\ref{tab:edgediff-metrics} presents summary statistics for the distribution over the 24 functions of evaluation differentials recorded on AN edges. This reflects networks constructed with 30 runs of the algorithms and $\epsilon$=0.00001; by \emph{evaluation differential} we mean the elapsed fitness evaluations at the destination node of the edge minus the elapsed evaluations at the source node of the edge. For each network, the median differential $md$ within it is recorded. The values in the table are the median and IQR value of $md$ across the 24 functions is reported. Each row captures data for a given algorithm and setting for $\beta$. The trend is that a higher $\beta$ leads to ANs with larger $md$. In the 2D case, {\de} has smaller $md$ values with lower IQR value when compared to {\cma}. In 10 dimensions, the two have similar $md$ values but there is a larger dispersion for {\de}. Random search {\rs} has, by far, the highest $md$ values of the three algorithms.

\begin{figure*}[tb!]
        \centering
\begin{subfigure}[b]{0.3\textwidth}  
            \centering 
            \includegraphics[trim=70 45 30 50,clip,width=\textwidth]{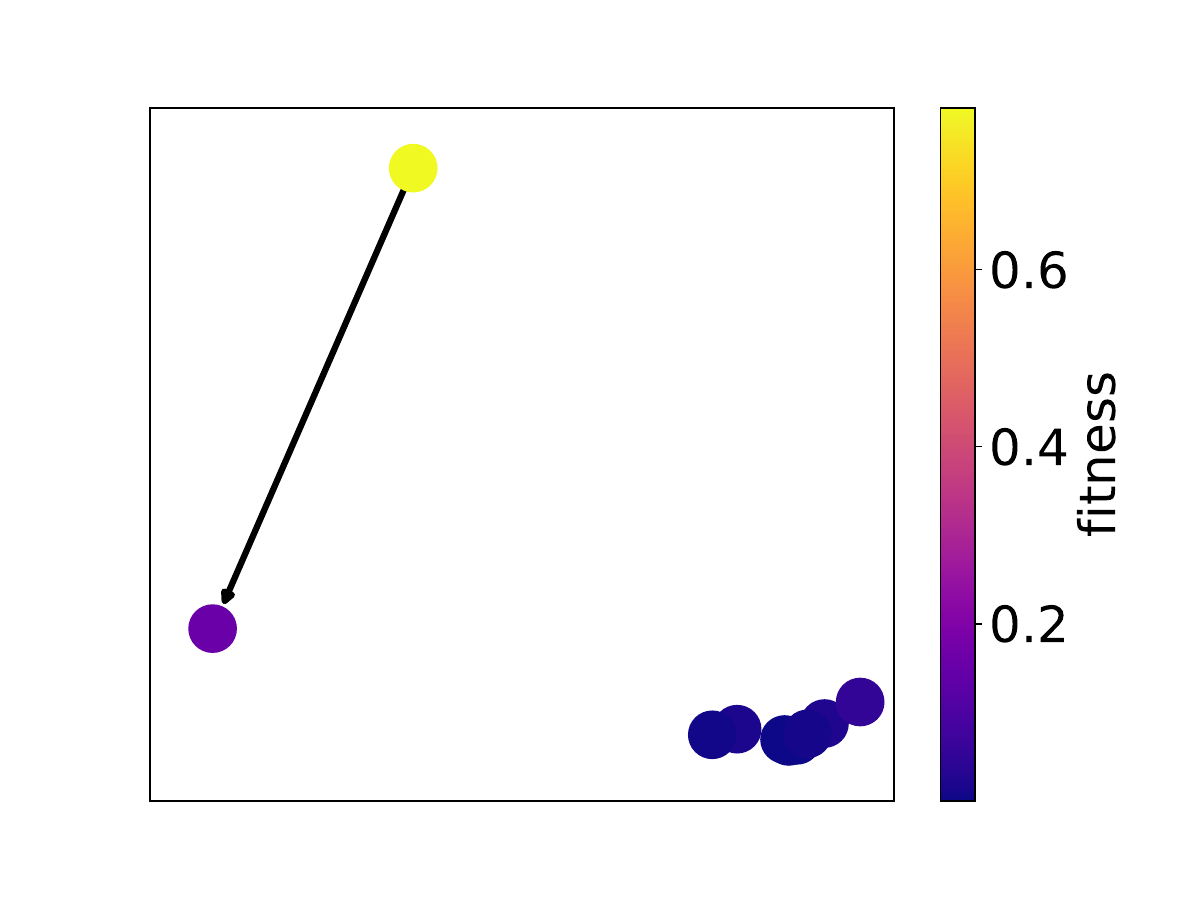}
            \caption{{\cma}}   
        \end{subfigure}
\begin{subfigure}[b]{0.3\textwidth}  
            \centering 
            \includegraphics[trim=70 45 30 50,clip,width=\textwidth]{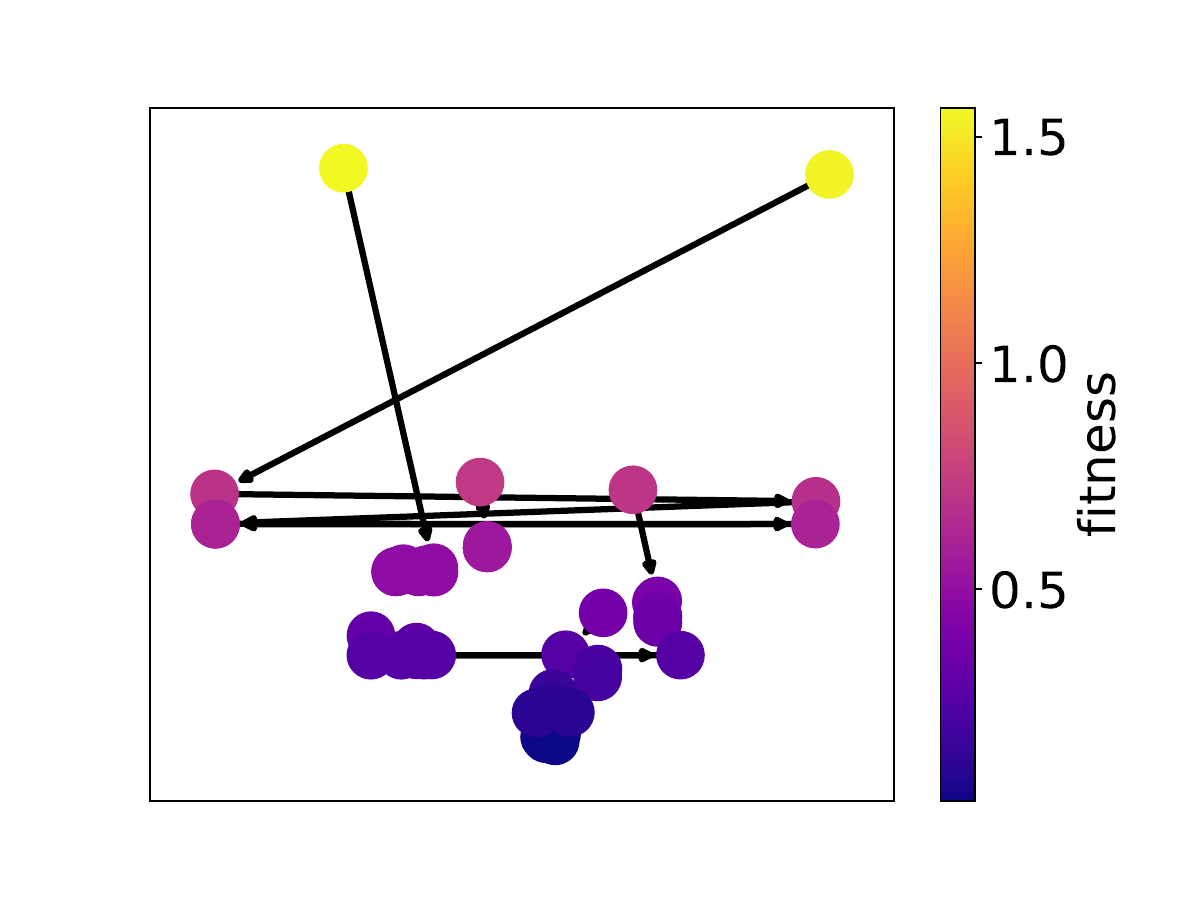}
            \caption{{\de}}   
        \end{subfigure}
        \begin{subfigure}[b]{0.3\textwidth}  
            \centering 
            \includegraphics[trim=70 45 30 50,clip,width=\textwidth]{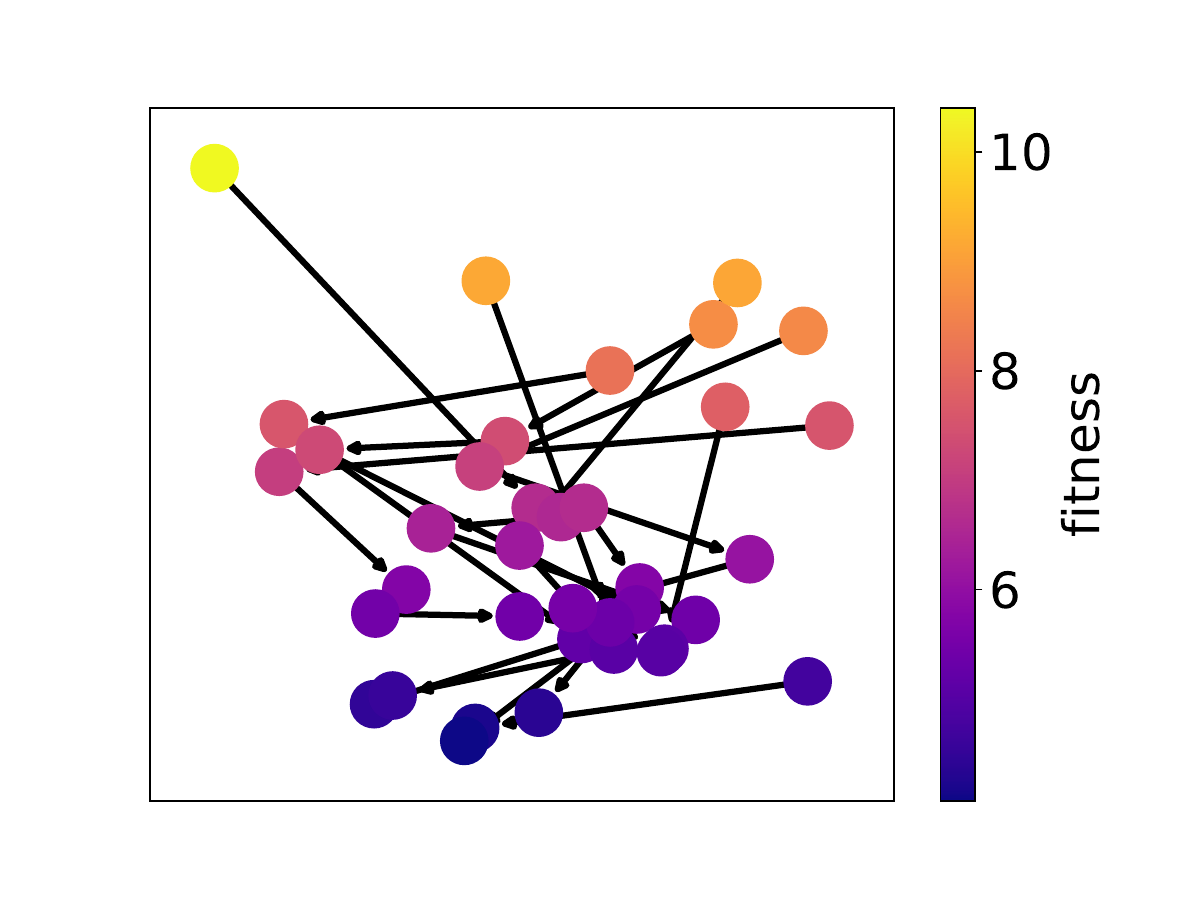}
            \caption{{\rs}}   
        \end{subfigure}
\caption{Attractor networks for 10D Schaffer function \emph{f17}, \(\beta\) = 320, $\epsilon$=\(10^{-5}\) for different algorithms}\label{fig:an-algorithms}
\end{figure*}

Figure~\ref{fig:an-algorithms} shows, for a fixed $\beta$ setting of 320, ANs for the three considered algorithms optimising 10-dimensional Schaffer function \emph{f17}. All three algorithms have ANs which look noticeably different. Random search {\rs} has the most dense (and most unstructured) network, and this is the trend across other functions as well (please see the supplemental material, where plots for all [function, algorithm, $\beta$, $\epsilon$] combinations are available~\cite{https://zenodo.org/records/14170241}). The {\de} network is smaller and a bit more structured: by following the arrows and the horizontal spacing, we can see that as search progresses there is movement towards a particular promising region of decision space. The {\cma} network is smaller still; we can see that at this setting of $\beta$=320 (that is, a stall of 32 generations), there is only a couple of sub-optimal attractors for this algorithm. 

\begin{figure*}[b!]
        \centering
  \begin{subfigure}[b]{0.35\textwidth}  
            \centering 
            \includegraphics[trim=10 10 5 5,clip,width=\textwidth]{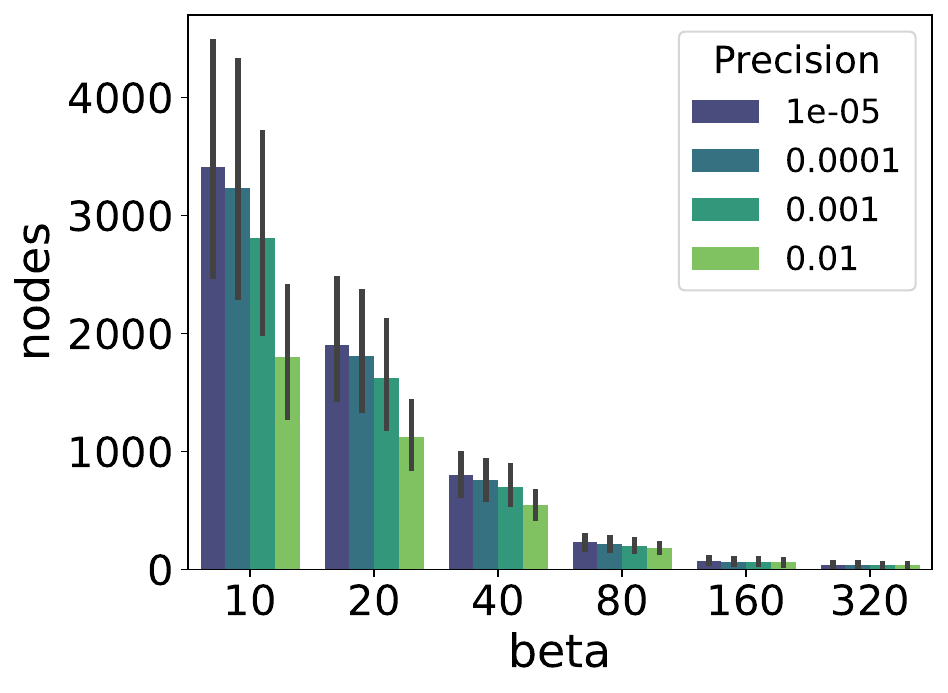}
            \caption{{\cma}}   
        \end{subfigure}
\begin{subfigure}[b]{0.35\textwidth}  
            \centering 
            \includegraphics[trim=10 10 5 5,clip,width=\textwidth]{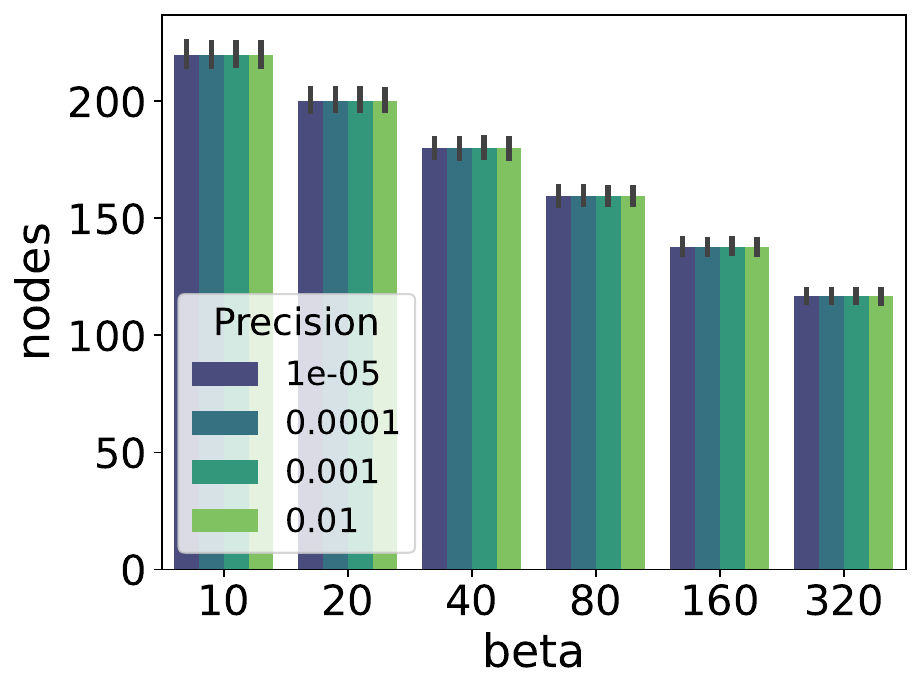}
            \caption{{\rs}}   
        \end{subfigure}
\caption{10D ANs [constructed with 30 runs] number of nodes with increasing $\beta$ and various $\epsilon$.}\label{fig:beta-versus-nodes}
\end{figure*}

Figure~\ref{fig:beta-versus-nodes} shows the change in network size (number of nodes) in ANs as $\beta$ increases and for various $\epsilon$, for {\cma} (left) and random search (right) on 10-dimensional functions. For each $\epsilon$ and $\beta$, there is a bar representing the median (over the 24 functions) and the variance is shown. Notice that within a coordinate precision level, nodes decrease approximately linearly with increasing $\beta$. In the case of {\cma}, a coarser $\epsilon$ leads to more nodes, and the variance decreases substantially with $\beta$. For random search, low $\beta$ leads to many fewer nodes than in the {\cma} counterparts, but there are still rather a lot of nodes at high $\beta$ (the decline in network size is much less dramatic in {\rs} than we see with {\cma}). 

\begin{figure*}[bt!]
        \centering
\includegraphics[trim=50 50 50 50,clip,width=0.995\textwidth]{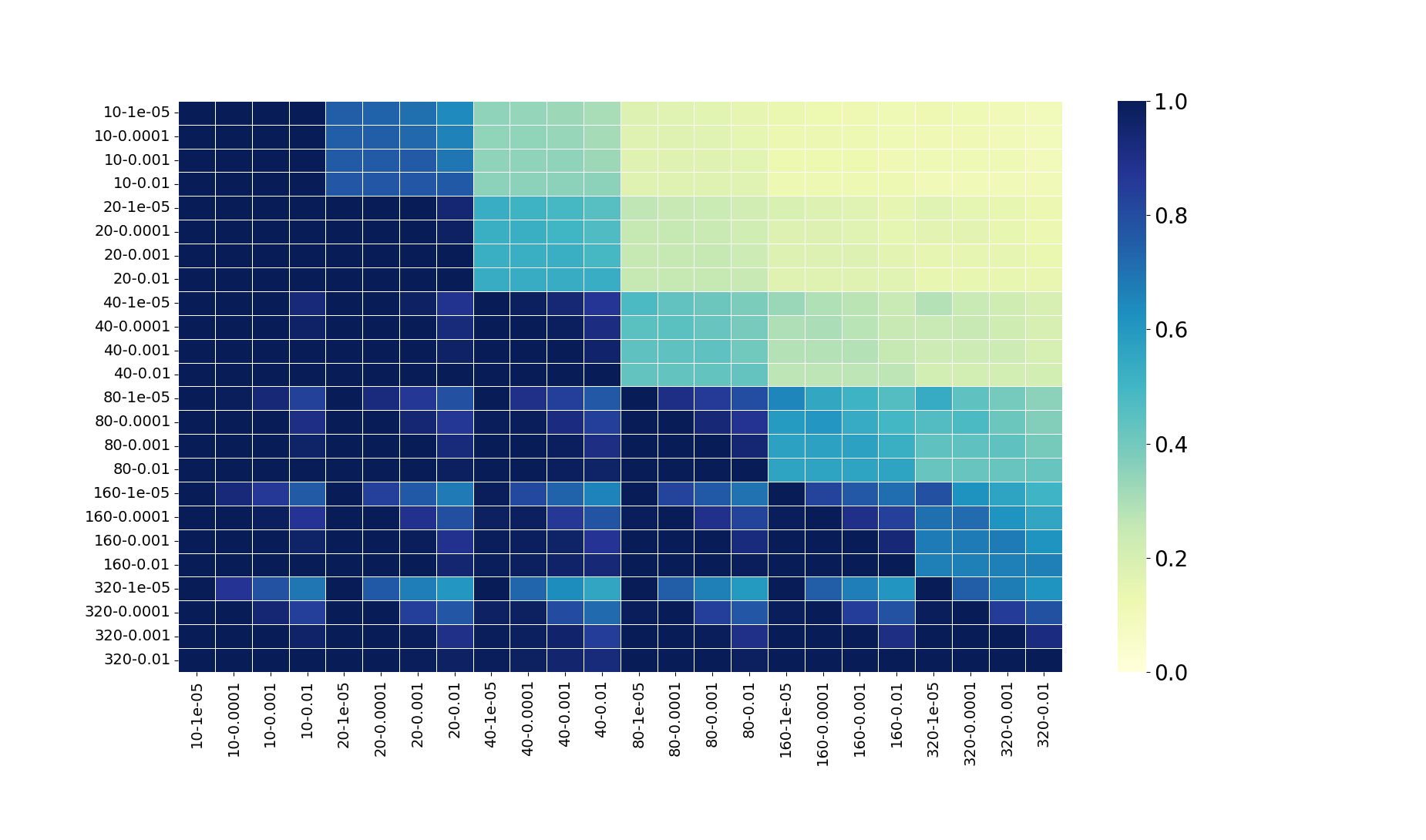}
\caption{The median proportion [over 24 functions] of node matches [i.e. fraction of mutual solutions] between {\cma} (2D) attractor networks with different network construction settings $\beta$ and $\epsilon$. In the plot labels, the integer part is the setting for $\beta$ and the float is for $\epsilon$}\label{fig:cma-combinations}
\end{figure*}

Figure~\ref{fig:cma-combinations} focuses on the extent to which attractor networks constructed using different configuration settings ($\beta$ and $\epsilon$) for the same functions have an overlap in the node locations for {\cma} on 2D functions. We are focussing more heavily on {\cma} than {\de}; this is because {\cma} tended to produce more interesting attractor networks. However, the same plot for {\de} can be found in the supplemental material. In Figure \ref{fig:cma-combinations}, pairs of AN configuration parameters are on each axis and the heat captures the median [across 24 functions] extent of mutual locations. For a given [vertical, horizontal] pairing, the heat value captures the fraction of the locations for the network type \emph{on the vertical axis} which are also present in the network type \emph{on the horizontal axis}. Squares with heat representing a value of 1.0 indicate that across the functions, that pair of differently-configured ANs have a complete node overlap. The purpose of this plot to to further understand the properties of ANs better. Looking at the plot, we notice from the dark blue regions that ANs with the same $\beta$ but different $\epsilon$ are often very similar. We can also observe that networks with a different (but similar) $\beta$ seem to frequently have some relation --- although this is to a lesser extent: in the region of 40-60\% of nodes matching. For the higher levels of $\beta$ (above 80), networks are also related in this way even to network counterparts which have a substantially different $\beta$ --- for example, pairs associated with the settings 80 and 320. In the top-right of the plot, we can notice that only a low proportion of nodes present in low-$\beta$ settings are also present with high $\beta$, which fits with intuition. 

\subsection{Network-based Landscape Models: a Comparison}

Table~\ref{tab:network-metrics} presents the median and IQR value [over 24 functions] for network nodes and edges with respect to LONs, STNs, and ANs with different $\beta$ configurations --- all of them constructed with a $\epsilon$=$10^{-5}$. Notice that networks associated with {\rs} are not present: this is because we noticed during STN construction that they were growing unmanageably large and deduced that they were unlikely to yield any sort of meaningful insight. Cell shading captures the proportion of the 24 functions where the associated network contains the global optimum; vibrant green means more often. 

Notice from the table that {\de} networks have fewer nodes and edges than their {\cma} counterparts. Comparing the LON metrics from the first row with other network types, we can see that on 2D functions, the LONs are of similar size to {\de} STNs; the {\cma} STNs are substantially larger than both of them. In fact, the 2D {\cma} STNs are larger even than the 10D networks. From looking at the raw data, we see that this is because the 2D {\cma} STNs are often comprised of many nodes with optimal or near-optimal fitness, but which are slightly different in decision space: at least $10^{-5}$ in one variable or more. There was a population size of six for this algorithm-dimension pair and the frequency of STN logging was every $k$ generations, so there could be a maximum of 37500 (1250 per run for 30 runs) STN nodes if no node was ever seen twice. On 10-dimensional functions, we observe that low-$\beta$ {\cma} ANs are larger than LONs, but that high-$\beta$ ANs are smaller than LONs. Notice also that for most {\de} ANs, the IQR value for the nodes and edges is actually the same. Through investigation, we found that this happens when the number of nodes is exactly 30 more than the number of edges in the network; the reason for this phenomenon occurring is attributable to the 30 separate runs used to construct the networks, and takes place when each of the runs terminates in a different location.
\begin{table*}[hb!]
     \caption{Number of network nodes and edges for networks [$\epsilon$=\(10^{-5}\)] constructed from 30 runs: median and IQR value [over 24 functions]. Cells are shaded according to how often [out of 24 functions] the global optimum is present in the network: more vibrant green is more often}
      \centering
        \resizebox{\textwidth}{!}{%
      \begin{tabular}{lcccc}
        \hline
        \multirow{2}{*}{\phantom{aaa}} & \multicolumn{2}{c}{2D} & \multicolumn{2}{c}{10D} \\
        \cline{2-5}
       model & nodes & edges & nodes & edges \\
         \hline
        LON &  \cellcolorbasedonvalue{96}{67.5 (143.25)} & \cellcolorbasedonvalue{96}{73 (164.5)} & \cellcolorbasedonvalue{54}{233 (521)} & \cellcolorbasedonvalue{54}{210 (520)} \\
        
        DE STN & \cellcolorbasedonvalue{33}{62 (94.5)} & \cellcolorbasedonvalue{33}{52.5 (94.75)} & \cellcolorbasedonvalue{0}{86.5 (52)} & \cellcolorbasedonvalue{0}{77 (56.5)} \\
        
       CMA STN &  \cellcolorbasedonvalue{88}{1367.5 (2166.75)} & \cellcolorbasedonvalue{88}{1585 (2268.25)} & \cellcolorbasedonvalue{50}{559 (1320)} & \cellcolorbasedonvalue{50}{579 (1361.5)} \\
       
       DE AN [$\beta$=40] & \cellcolorbasedonvalue{33}{46.5 (8.25)} & \cellcolorbasedonvalue{33}{16.5 (8.25)} & \cellcolorbasedonvalue{4}{416.5 (141.25)} & \cellcolorbasedonvalue{4}{389 (140.25)} \\
       DE AN [$\beta$=80] & \cellcolorbasedonvalue{33}{31 (3)} & \cellcolorbasedonvalue{33}{1.5 (3)} & \cellcolorbasedonvalue{4}{138 (223.25)} & \cellcolorbasedonvalue{4}{108 (223.25)} \\

       CMA AN [$\beta$=40] & \cellcolorbasedonvalue{96}{104 (41.5)} & \cellcolorbasedonvalue{96}{96 (39.5)} & \cellcolorbasedonvalue{50}{860 (558.25)} & \cellcolorbasedonvalue{50}{830 (570)} \\
       CMA AN [$\beta$=80] & \cellcolorbasedonvalue{96}{44.5 (25.5)} & \cellcolorbasedonvalue{96}{29 (20.25)} & \cellcolorbasedonvalue{50}{204.5 (216.25)} & \cellcolorbasedonvalue{50}{196 (195.25)} \\

       \hline
    \end{tabular}}
\end{table*}

\begin{figure*}[!t]
        \centering
    \begin{subfigure}[b]{0.24\textwidth}  
        \centering 
        \includegraphics[trim=70 45 30 50,clip,width=\textwidth]{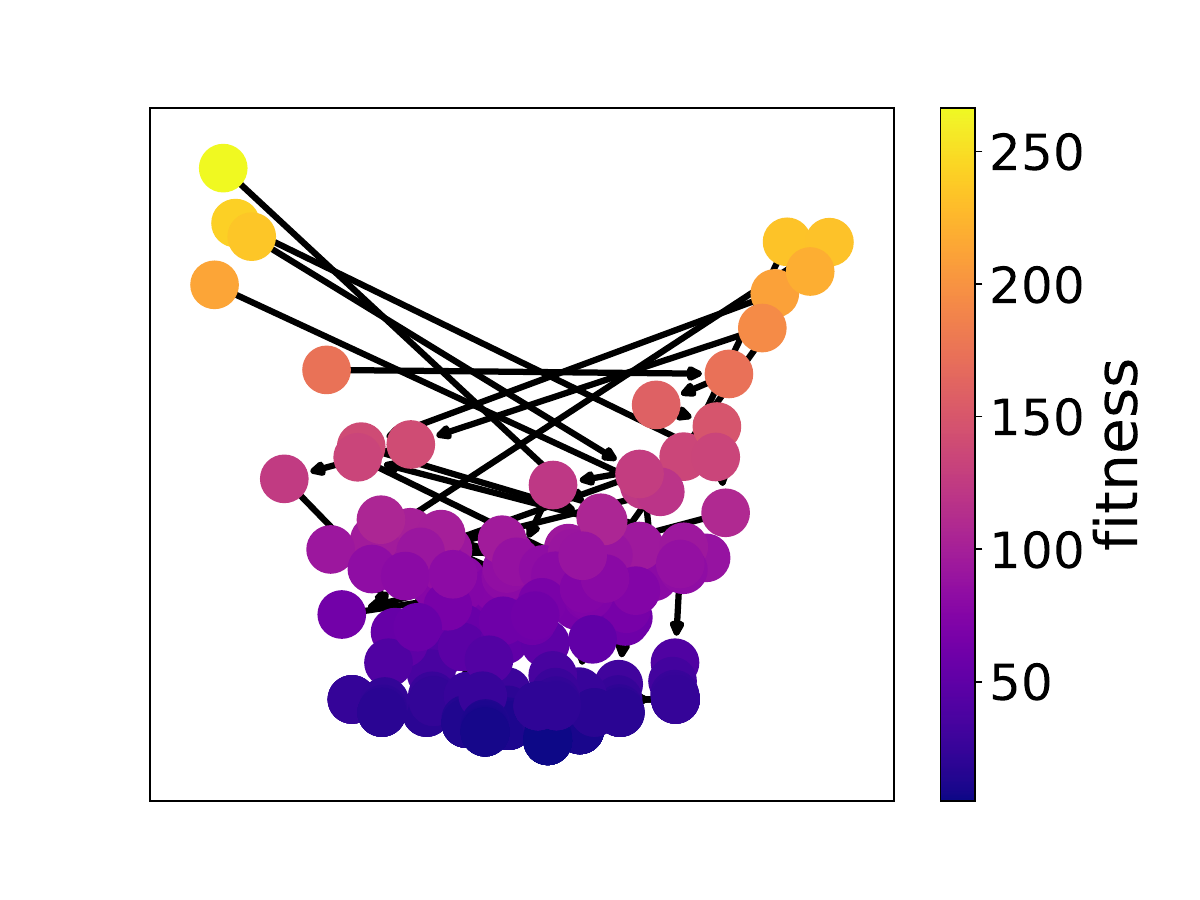}
        \caption{{\cma} STN}\label{fig:stn}
    \end{subfigure}
    \begin{subfigure}[b]{0.24\textwidth}  
        \centering 
        \includegraphics[trim=70 45 30 50,clip,width=\textwidth]{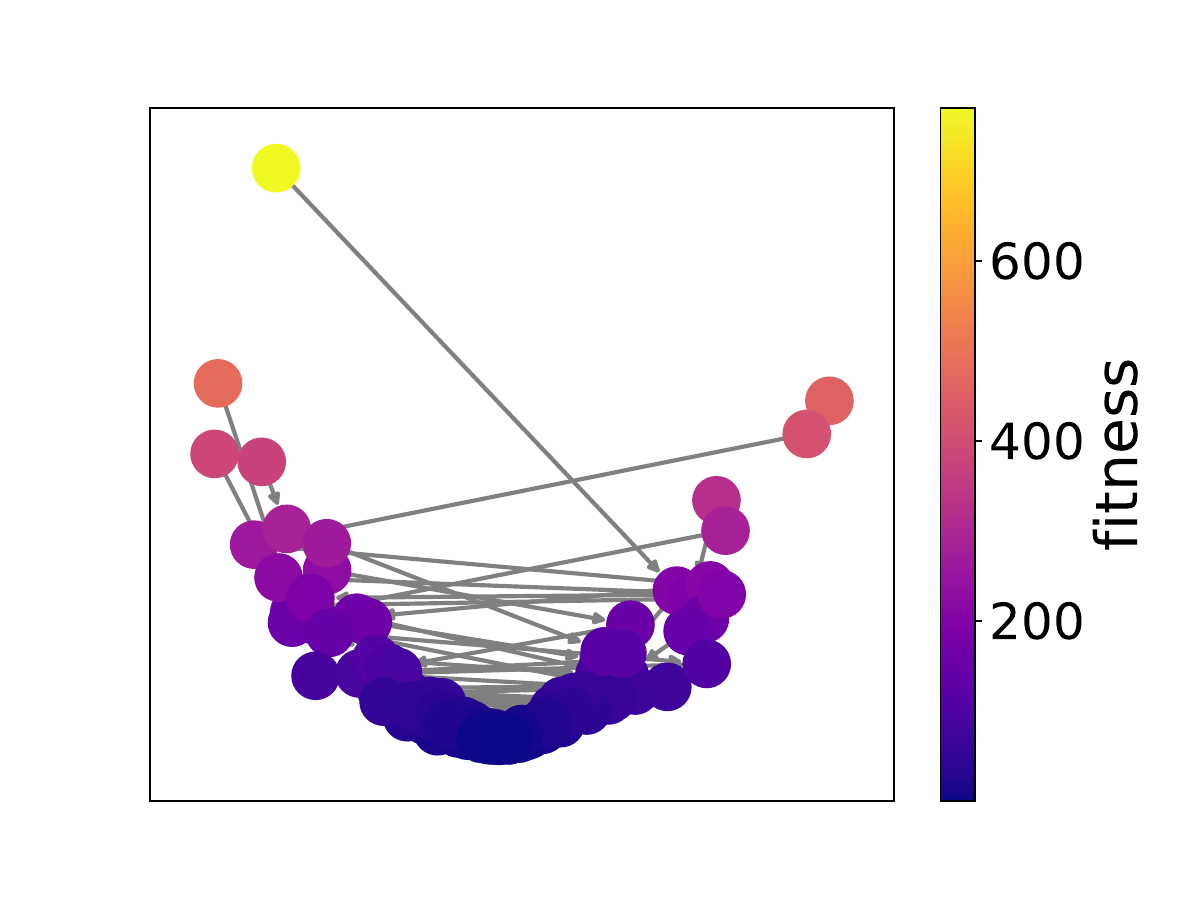}
        \caption{{\mbh} LON} \label{fig:lon}
    \end{subfigure}
    \begin{subfigure}[b]{0.24\textwidth}  
        \centering 
        \includegraphics[trim=70 45 30 50,clip,width=\textwidth]{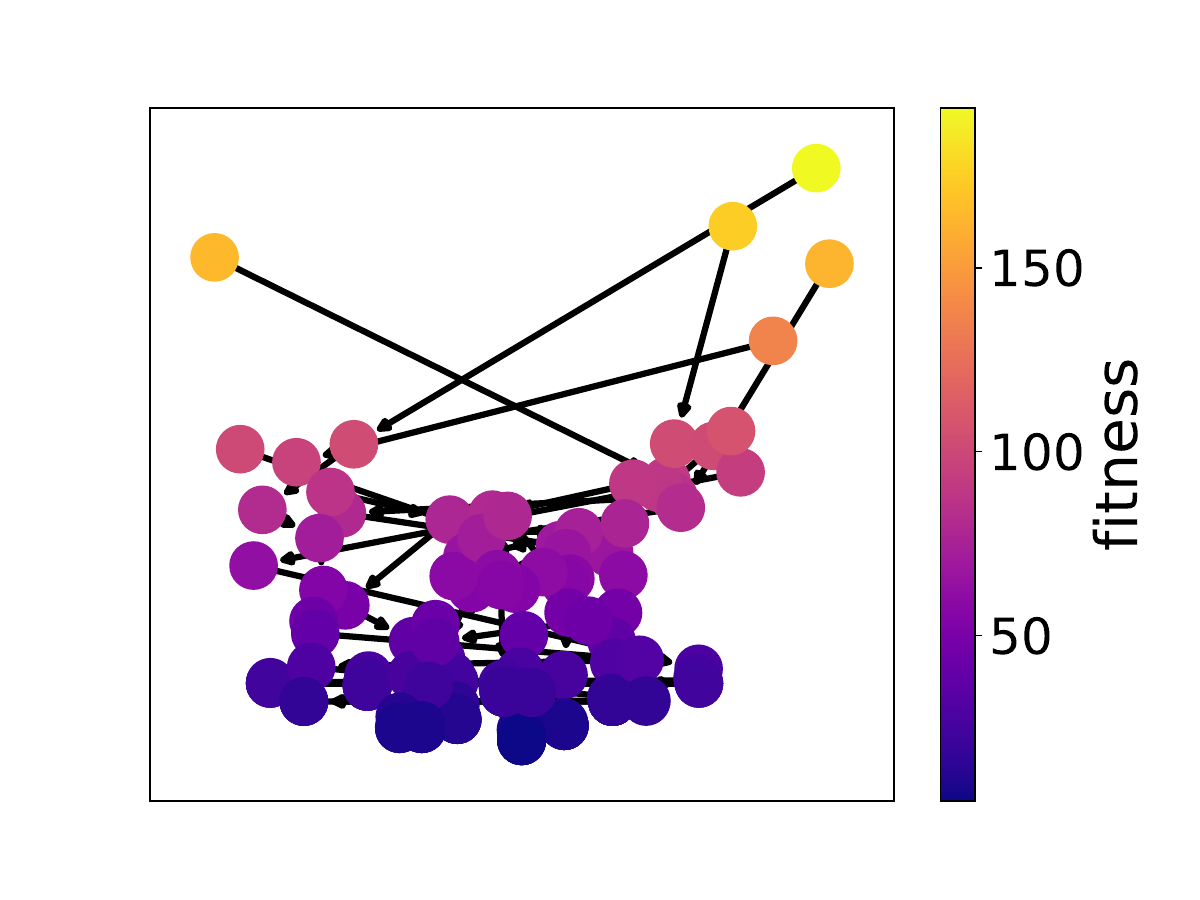}
        \caption{{\cma}-AN \(\beta\)=40} \label{fig:an-low}
    \end{subfigure}
    \begin{subfigure}[b]{0.24\textwidth}  
        \centering 
        \includegraphics[trim=70 45 30 50,clip,width=\textwidth]{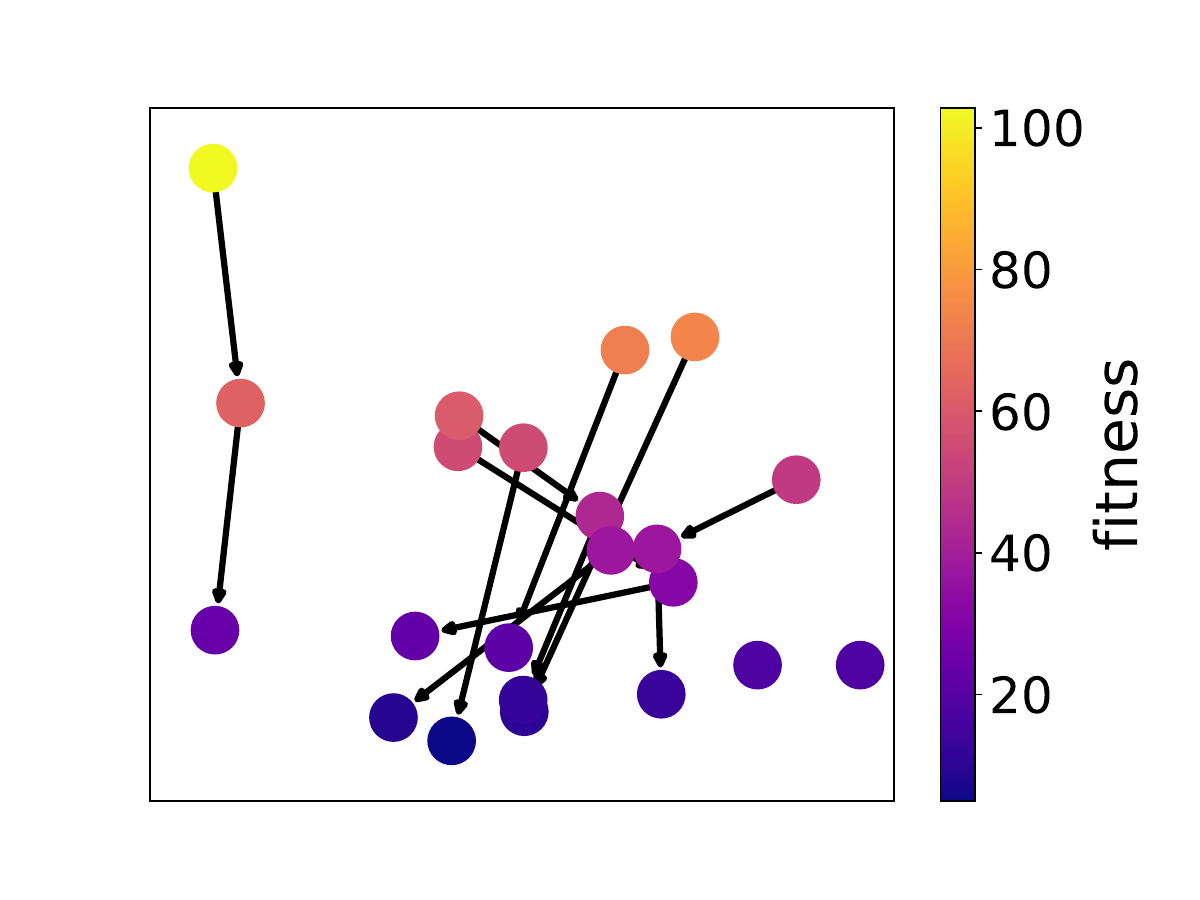}
        \caption{{\cma}-AN \(\beta\)=160} \label{fig:an-high}
    \end{subfigure}
\caption{Networks [$\epsilon$=\(10^{-5}\)] for 10D Rosenbrock function \emph{f3}; the STN and AN show {\cma} behaviour, while the LON is constructed according to {\mbh} (that is the convention)}\label{fig:three-network-types}
\end{figure*}

Figure~\ref{fig:three-network-types} is an illustrative comparison between different network model types: LONs, {\cma} STNs, and {\cma} ANs [with two $\beta$ settings] are shown. Note that visualisations for the other problem instances are available in the supplemental material~\cite{https://zenodo.org/records/14170241}. In the case of Figure~\ref{fig:three-network-types}, all plots reflect algorithms running on 10-dimensional Rastrign function \emph{f3}. 

Surveying the figure, we notice that the STN, the LON, and the low $\beta$ AN seem to reveal the overall structure of how algorithms move on the problem (forming a single-funnel shape). In the case of the LON, each node had been obtained through local search. In the case of the STN (Figure~\ref{fig:stn}), nodes are not necessarily attractors. The low $\beta$ AN resembles the STN. For the high $\beta$ AN, there is a focus on exclusively attractors a.k.a. "local optima" for {\cma}. While each node in the LON (\ref{fig:lon}) is a local optimum for {\mbh}, every node in Figure~\ref{fig:an-high} is an attractor for {\cma}. We can see that the latter is the sparsest of the four, and allows us to see a visualisation of search trajectories with a \textbf{temporal} component than none of the other models can show: we know that {\cma} stalled at each of these locations for at least 160 evaluations [16 generations]. Although an STN or LON can convey how often search passed through locations, they do not convey \textbf{how long} it spent there. The STN visualised here is rather crowded and it would not be possible to know which nodes are temporal attractors. While the LON does show local optima, there are two limitations: a) the temporal aspect is not captured and b) the local optima are with respect to {\mbh}, rather than the algorithm under study (here: {\cma}). 

\begin{figure*}[b!]
        \centering
   \begin{subfigure}[b]{0.4\textwidth}
            \centering
            \includegraphics[trim=25 30 25 25,clip,width=\textwidth]{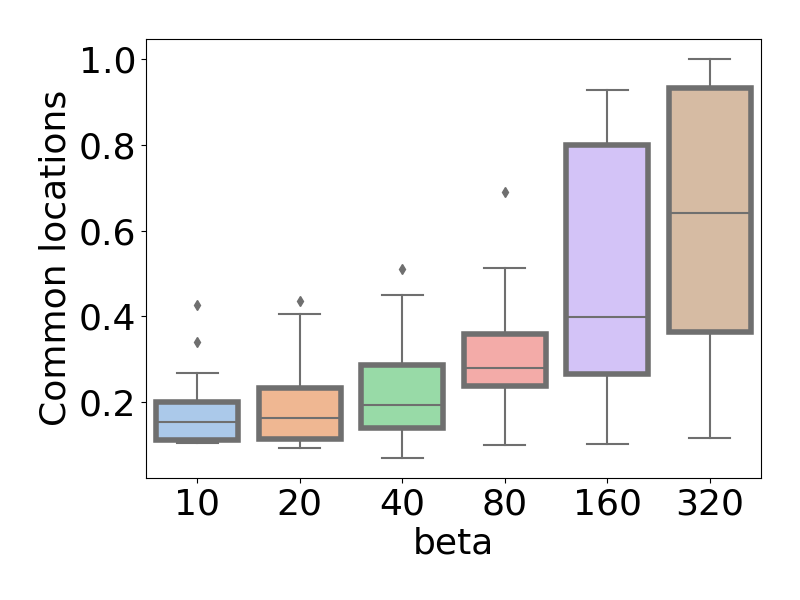}
            \caption{STN}
            \label{fig:stn-matches}
        \end{subfigure}
        \hspace{1em}
 \begin{subfigure}[b]{0.4\textwidth}
            \centering
            \includegraphics[trim=25 30 25 25,clip,width=\textwidth]{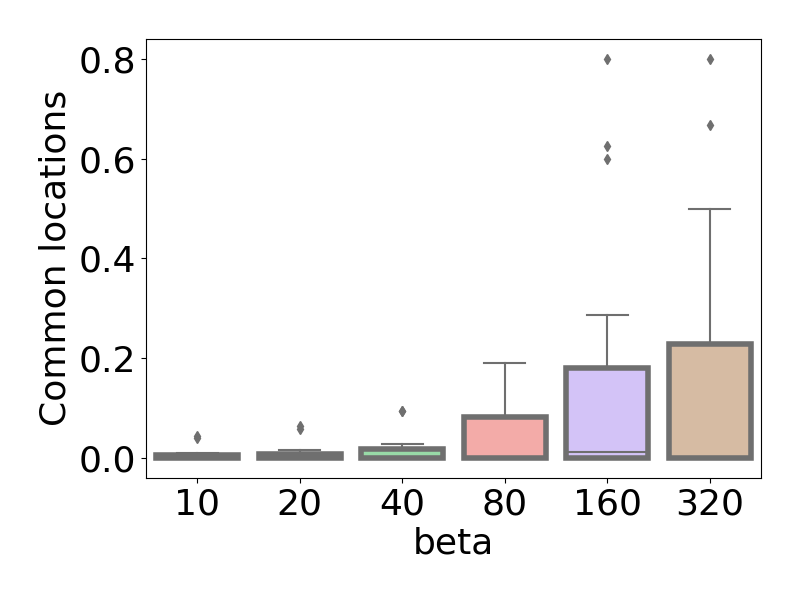}
            \caption{LON}
            \label{fig:lon-matches}
        \end{subfigure}
\caption{{\cma} 10D proportion of matching locations between AN and other network models for all 24 functions [networks built with 30 runs and $\epsilon$=\(10^{-2}\)] across increasing AN $\beta$; represented as a proportion of AN size.} \label{fig:match-locs}
\end{figure*}

The two plots in Figure~\ref{fig:match-locs} show the proportion of {\cma} AN nodes which are also present in their corresponding STN (\ref{fig:stn-matches}) and LON (\ref{fig:lon-matches}) respectively. The networks were constructed with 30 runs and $\epsilon$=0.01 is used. Various $\beta$ settings for the ANs are considered. We can observe that as $\beta$ increases, the proportion of AN nodes increases as well. This is because there are less AN nodes and the ones which \emph{are} still present are strong attractors; it makes sense that these nodes would also be present in the STN and LON. As $\beta$ increases, though, the variance increases --- particularly in the case of the STN matching in Figure~\ref{fig:stn-matches}. This implies that the degree of matching as it relates to high $\beta$ ANs may depend on the nature of the particular function. The LON overlaps are less substantial than the STN overlaps; this makes sense, because the LON is constructed using {\mbh}, while the STNs and ANs are constructed using {\cma} (in this case). It seems that {\mbh} and for {\cma} traverse many \textit{different} parts of the search space and for {\cma} attractors do \textit{not} seem to be equivalent to LON local optima in most cases.

\section{Limitations and Outlook}\label{limitations}

While attractor networks (ANs) offer an interesting way for understanding algorithmic stalling behaviour across different optimization landscapes, several limitations should be taken into consideration. Firstly, the construction and interpretation of ANs depend significantly on both the algorithm and the optimisation problem. The AN assumes a single population or individual traversing the search space. Methods that introduce diversity maintenance, such as niching and quality diversity algorithms, \textit{may} result in misleading or overly complex ANs. For these approaches with multiple subpopulations, attractors might not represent stalling but rather indicate ongoing exploration within different niches. Therefore, alternative visualizations or constructions would be required to accommodate algorithms that are inherently multimodal or explicitly diversity-promoting. Another consideration is the sample dependency in AN construction. For problems with highly multimodal landscapes or higher dimensionality, the attractor network structure \textit{may} vary substantially across different runs. This necessitates additional runs to capture a more comprehensive network structure and to ensure robustness in AN-derived insights. The use of a fixed evaluation threshold for stalling also presents limitations in long-running or fine-tuned searches, where stagnation detection would benefit from adaptive thresholds. A future direction could be to explore an increasing window for evaluation that adapts to the  progression of the search. Despite these limitations, the specificity of ANs to individual algorithms also gives an advantage. It allows for algorithm-level comparisons that go beyond performance metrics alone. This could enable a more nuanced understanding of algorithm dynamics, particularly in contexts where standard metrics are insufficient to capture structural differences. Looking forward, expanding the application of ANs across diverse algorithm classes and optimization scenarios could provide a better method to analyse algorithm behaviour and problem landscapes. The AN framework can be adapted for compatibility with multimodal algorithms or for discrete problem spaces. Additionally, future studies could explore hybrid network models that integrate insights from both traditional local optima networks and search trajectory networks.

\section{Conclusions}%
\label{sec:conclusions}%

In this work, we formalised and put forward the notion of attractor networks (ANs) as a novel framework for analysing the stalling behaviour of optimization algorithms. By focusing on attractor points --- locations in the search space where algorithms experience prolonged stagnation --- ANs provide a lens for examining algorithm dynamics beyond the reach of traditional local optima networks (LONs) and search trajectory networks (STNs). Unlike LONs, which are limited to hill-climbing algorithms, and STNs, which do not typically emphasise intermediate stalling behaviour, ANs facilitate a structured view of algorithm trajectories for any optimisation approach --- including those that don't use local search. Through systematic analysis across 24 BBOB functions, we demonstrated that ANs reveal meaningful contrasts in how CMA-ES, differential evolution, and random search engage with search spaces. We show that ANs can give insights into intermediate attractors where alternative network-based models of algorithm behaviour would not.
The AN model's flexibility in capturing unique behavioural characteristics presents a valuable direction for comparative analysis, enabling insights into algorithm-specific stalling and convergence patterns. Future studies may adapt the AN approach to accommodate more complex multimodal search strategies, broadening the applicability of ANs within optimization research. Code and data for this paper are available in a Zenodo repository \cite{https://zenodo.org/records/14170241}.
%
\bibliographystyle{splncs03}%
\bibliography{an}%
\end{document}


\title{Entropy, Search Trajectories, and Explainability \\ 
for Frequency Fitness Assignment --- supplemental material}%

\author{Sarah L. Thomson$^1$ \and Gabriela Ochoa$^2$ \and 
Daan van den Berg$^3$ \and \\ Tianyu Liang$^4$ \and Thomas Weise$^4$}
\authorrunning{Thomson et al.}
%
\institute{Edinburgh Napier University, UK
\email{s.thomson4@napier.ac.uk}
\and University of Stirling, UK
\email{gabriela.ochoa@cs.stir.ac.uk}
\and Vrije Universiteit, Netherlands
\email{daan@yamasan.nl}
\and Hefei University, China
\email{liangty@stu.hfuu.edu.cn,tweise@ustc.edu.cn}
}
\maketitle

\section{Limitations of layout QAP}

\begin{enumerate}
    \item Nearest neighbors are attempted to be put closely together. However: very different solutions are not necessarily far apart (this condition maybe cannot be represented in a QAP). But since we map a very high-dimensional space to 1-d, trying to enforce that far-away solutions are also far-away while enforcing that near solutions are near would lead to a very constrained and hard problem. What we can say is that if two solutions are very similar, then they should not be far away in the graph (if the QAP is solved well).
    \item However: the swap distance is between 1 (one swap is needed to get from solution A to solution B) and \scale. It can never be more than \scale. Also, each solution can reach $0.5\scale(\scale-1)$ other solution with 1 swap. These two facts together mean that many solutions will have the same distance. They then share the same distance rank in the layout QAP. They then have the same flow between each other. If we have thousands and thousands of solutions, it may be that for each solution, there could be dozens of solutions with distance~1, i.e., nearest neighbors, in the set of captured solutions. But we can only pick two of them to be the actual nearest neighbors on the x-coordinates of a given solution, since there are only 2 nearest coordinates on a 1-d axis. In the worst case, if half of the solutions are nearest neighbors of each other, the graph would look vastly random. However, this would likely be the case for any visualization in such a situation.
    \item The helper QAP is solved with OHC {\dots} which was the worst algorithm in our study. This means the results are very likely local optima. In the future, we will consider \objSA{}.
    \item On a 1-D x-axis, each solution has either 1 nearest neighbor (if it is on the left or right end of the axis) or exactly two nearest neighbors (otherwise). However, in our permutation space, each solution has has $0.5\scale(\scale-1)$ nearest neighbors. Now maybe our set of collected solutions does not contain all of them, maybe even just one. Also, in 1-D, each solution has either 1 or 2 second-nearest neighbors, either 1 or 2 third-nearest neighbors, and so on. But in our permutation space, their may be cubic or quartic many such neighbors. The space potentially is much more high-dimensional. It is not clear whether it can even be mapped correctly to 1 dimension. So, the layout QAP is just a coarse heuristic.
\end{enumerate}

\section{Search trajectory networks}
In this Section, the search trajectory networks (STNs) for the remaining 19 QAP instances (\textbf{tai27e02-20}) are shown. 

 \begin{figure}[h]
        \centering
        \begin{subfigure}[b]{0.49\textwidth}
            \centering
            \includegraphics[width=\textwidth]{supplemental/tai27e02-OHC.pdf}
            \caption{Objective-steered hill-climbing \objHC}
            \label{fig:ohc-stn}
        \end{subfigure}
        \hfill
        \begin{subfigure}[b]{0.49\textwidth}
            \centering
            \includegraphics[width=\textwidth]{supplemental/tai27e02-FHC.pdf}
            \caption{FFA-based hill-climbing \ffaHC}
            \label{fig:fhc-stn}
        \end{subfigure}
        \vskip\baselineskip
        \begin{subfigure}[b]{0.49\textwidth}
            \centering
            \includegraphics[width=\textwidth]{supplemental/tai27e02-OSA.pdf}
            \caption{Objective-steered SA \objSA}
            \label{fig:osa-stn}
        \end{subfigure}
        \hfill
        \begin{subfigure}[b]{0.49\textwidth}
            \centering
            \includegraphics[width=\textwidth]{supplemental/tai27e02-FSA.pdf}
            \caption{FFA-based simulated annealing \ffaSA}
            \label{fig:fsa-stn}
        \end{subfigure}
        \caption{Search trajectory networks for the QAP instance~\textbf{tai27e02}. For the purpose of visual clarity, only five runs are shown per plot. Each run is a different colour. Fitness is on the $y$-axis and solutions are placed in an ordering along the horizontal axis which is obtained from optimising their placement according to the relative permutation swap distances}
            \label{fig:stns}
    \end{figure}%

 \begin{figure}[h]
        \centering
        \begin{subfigure}[b]{0.49\textwidth}
            \centering
            \includegraphics[width=\textwidth]{supplemental/tai27e03-OHC.pdf}
            \caption{Objective-steered hill-climbing \objHC}
            \label{fig:ohc-stn}
        \end{subfigure}
        \hfill
        \begin{subfigure}[b]{0.49\textwidth}
            \centering
            \includegraphics[width=\textwidth]{supplemental/tai27e03-FHC.pdf}
            \caption{FFA-based hill-climbing \ffaHC}
            \label{fig:fhc-stn}
        \end{subfigure}
        \vskip\baselineskip
        \begin{subfigure}[b]{0.49\textwidth}
            \centering
            \includegraphics[width=\textwidth]{supplemental/tai27e03-OSA.pdf}
            \caption{Objective-steered SA \objSA}
            \label{fig:osa-stn}
        \end{subfigure}
        \hfill
        \begin{subfigure}[b]{0.49\textwidth}
            \centering
            \includegraphics[width=\textwidth]{supplemental/tai27e03-FSA.pdf}
            \caption{FFA-based simulated annealing \ffaSA}
            \label{fig:fsa-stn}
        \end{subfigure}
        \caption{Search trajectory networks for the QAP instance~\textbf{tai27e03}. For the purpose of visual clarity, only five runs are shown per plot. Each run is a different colour. Fitness is on the $y$-axis and solutions are placed in an ordering along the horizontal axis which is obtained from optimising their placement according to the relative permutation swap distances}
            \label{fig:stns}
    \end{figure}%

     \begin{figure}[h]
        \centering
        \begin{subfigure}[b]{0.49\textwidth}
            \centering
            \includegraphics[width=\textwidth]{supplemental/tai27e04-OHC.pdf}
            \caption{Objective-steered hill-climbing \objHC}
            \label{fig:ohc-stn}
        \end{subfigure}
        \hfill
        \begin{subfigure}[b]{0.49\textwidth}
            \centering
            \includegraphics[width=\textwidth]{supplemental/tai27e04-FHC.pdf}
            \caption{FFA-based hill-climbing \ffaHC}
            \label{fig:fhc-stn}
        \end{subfigure}
        \vskip\baselineskip
        \begin{subfigure}[b]{0.49\textwidth}
            \centering
            \includegraphics[width=\textwidth]{supplemental/tai27e04-OSA.pdf}
            \caption{Objective-steered SA \objSA}
            \label{fig:osa-stn}
        \end{subfigure}
        \hfill
        \begin{subfigure}[b]{0.49\textwidth}
            \centering
            \includegraphics[width=\textwidth]{supplemental/tai27e04-FSA.pdf}
            \caption{FFA-based simulated annealing \ffaSA}
            \label{fig:fsa-stn}
        \end{subfigure}
        \caption{Search trajectory networks for the QAP instance~\textbf{tai27e04}. For the purpose of visual clarity, only five runs are shown per plot. Each run is a different colour. Fitness is on the $y$-axis and solutions are placed in an ordering along the horizontal axis which is obtained from optimising their placement according to the relative permutation swap distances}
            \label{fig:stns}
    \end{figure}%

 \begin{figure}[h]
        \centering
        \begin{subfigure}[b]{0.49\textwidth}
            \centering
            \includegraphics[width=\textwidth]{supplemental/tai27e05-OHC.pdf}
            \caption{Objective-steered hill-climbing \objHC}
            \label{fig:ohc-stn}
        \end{subfigure}
        \hfill
        \begin{subfigure}[b]{0.49\textwidth}
            \centering
            \includegraphics[width=\textwidth]{supplemental/tai27e05-FHC.pdf}
            \caption{FFA-based hill-climbing \ffaHC}
            \label{fig:fhc-stn}
        \end{subfigure}
        \vskip\baselineskip
        \begin{subfigure}[b]{0.49\textwidth}
            \centering
            \includegraphics[width=\textwidth]{supplemental/tai27e05-OSA.pdf}
            \caption{Objective-steered SA \objSA}
            \label{fig:osa-stn}
        \end{subfigure}
        \hfill
        \begin{subfigure}[b]{0.49\textwidth}
            \centering
            \includegraphics[width=\textwidth]{supplemental/tai27e05-FSA.pdf}
            \caption{FFA-based simulated annealing \ffaSA}
            \label{fig:fsa-stn}
        \end{subfigure}
        \caption{Search trajectory networks for the QAP instance~\textbf{tai27e05}. For the purpose of visual clarity, only five runs are shown per plot. Each run is a different colour. Fitness is on the $y$-axis and solutions are placed in an ordering along the horizontal axis which is obtained from optimising their placement according to the relative permutation swap distances}
            \label{fig:stns}
    \end{figure}%

 \begin{figure}[h]
        \centering
        \begin{subfigure}[b]{0.49\textwidth}
            \centering
            \includegraphics[width=\textwidth]{supplemental/tai27e06-OHC.pdf}
            \caption{Objective-steered hill-climbing \objHC}
            \label{fig:ohc-stn}
        \end{subfigure}
        \hfill
        \begin{subfigure}[b]{0.49\textwidth}
            \centering
            \includegraphics[width=\textwidth]{supplemental/tai27e06-FHC.pdf}
            \caption{FFA-based hill-climbing \ffaHC}
            \label{fig:fhc-stn}
        \end{subfigure}
        \vskip\baselineskip
        \begin{subfigure}[b]{0.49\textwidth}
            \centering
            \includegraphics[width=\textwidth]{supplemental/tai27e06-OSA.pdf}
            \caption{Objective-steered SA \objSA}
            \label{fig:osa-stn}
        \end{subfigure}
        \hfill
        \begin{subfigure}[b]{0.49\textwidth}
            \centering
            \includegraphics[width=\textwidth]{supplemental/tai27e06-FSA.pdf}
            \caption{FFA-based simulated annealing \ffaSA}
            \label{fig:fsa-stn}
        \end{subfigure}
        \caption{Search trajectory networks for the QAP instance~\textbf{tai27e06}. For the purpose of visual clarity, only five runs are shown per plot. Each run is a different colour. Fitness is on the $y$-axis and solutions are placed in an ordering along the horizontal axis which is obtained from optimising their placement according to the relative permutation swap distances}
            \label{fig:stns}
    \end{figure}%

 \begin{figure}[h]
        \centering
        \begin{subfigure}[b]{0.49\textwidth}
            \centering
            \includegraphics[width=\textwidth]{supplemental/tai27e07-OHC.pdf}
            \caption{Objective-steered hill-climbing \objHC}
            \label{fig:ohc-stn}
        \end{subfigure}
        \hfill
        \begin{subfigure}[b]{0.49\textwidth}
            \centering
            \includegraphics[width=\textwidth]{supplemental/tai27e07-FHC.pdf}
            \caption{FFA-based hill-climbing \ffaHC}
            \label{fig:fhc-stn}
        \end{subfigure}
        \vskip\baselineskip
        \begin{subfigure}[b]{0.49\textwidth}
            \centering
            \includegraphics[width=\textwidth]{supplemental/tai27e07-OSA.pdf}
            \caption{Objective-steered SA \objSA}
            \label{fig:osa-stn}
        \end{subfigure}
        \hfill
        \begin{subfigure}[b]{0.49\textwidth}
            \centering
            \includegraphics[width=\textwidth]{supplemental/tai27e07-FSA.pdf}
            \caption{FFA-based simulated annealing \ffaSA}
            \label{fig:fsa-stn}
        \end{subfigure}
        \caption{Search trajectory networks for the QAP instance~\textbf{tai27e07}. For the purpose of visual clarity, only five runs are shown per plot. Each run is a different colour. Fitness is on the $y$-axis and solutions are placed in an ordering along the horizontal axis which is obtained from optimising their placement according to the relative permutation swap distances}
            \label{fig:stns}
    \end{figure}%

 \begin{figure}[h]
        \centering
        \begin{subfigure}[b]{0.49\textwidth}
            \centering
            \includegraphics[width=\textwidth]{supplemental/tai27e08-OHC.pdf}
            \caption{Objective-steered hill-climbing \objHC}
            \label{fig:ohc-stn}
        \end{subfigure}
        \hfill
        \begin{subfigure}[b]{0.49\textwidth}
            \centering
            \includegraphics[width=\textwidth]{supplemental/tai27e08-FHC.pdf}
            \caption{FFA-based hill-climbing \ffaHC}
            \label{fig:fhc-stn}
        \end{subfigure}
        \vskip\baselineskip
        \begin{subfigure}[b]{0.49\textwidth}
            \centering
            \includegraphics[width=\textwidth]{supplemental/tai27e08-OSA.pdf}
            \caption{Objective-steered SA \objSA}
            \label{fig:osa-stn}
        \end{subfigure}
        \hfill
        \begin{subfigure}[b]{0.49\textwidth}
            \centering
            \includegraphics[width=\textwidth]{supplemental/tai27e08-FSA.pdf}
            \caption{FFA-based simulated annealing \ffaSA}
            \label{fig:fsa-stn}
        \end{subfigure}
        \caption{Search trajectory networks for the QAP instance~\textbf{tai27e08}. For the purpose of visual clarity, only five runs are shown per plot. Each run is a different colour. Fitness is on the $y$-axis and solutions are placed in an ordering along the horizontal axis which is obtained from optimising their placement according to the relative permutation swap distances}
            \label{fig:stns}
    \end{figure}%

 \begin{figure}[h]
        \centering
        \begin{subfigure}[b]{0.49\textwidth}
            \centering
            \includegraphics[width=\textwidth]{supplemental/tai27e09-OHC.pdf}
            \caption{Objective-steered hill-climbing \objHC}
            \label{fig:ohc-stn}
        \end{subfigure}
        \hfill
        \begin{subfigure}[b]{0.49\textwidth}
            \centering
            \includegraphics[width=\textwidth]{supplemental/tai27e09-FHC.pdf}
            \caption{FFA-based hill-climbing \ffaHC}
            \label{fig:fhc-stn}
        \end{subfigure}
        \vskip\baselineskip
        \begin{subfigure}[b]{0.49\textwidth}
            \centering
            \includegraphics[width=\textwidth]{supplemental/tai27e09-OSA.pdf}
            \caption{Objective-steered SA \objSA}
            \label{fig:osa-stn}
        \end{subfigure}
        \hfill
        \begin{subfigure}[b]{0.49\textwidth}
            \centering
            \includegraphics[width=\textwidth]{supplemental/tai27e09-FSA.pdf}
            \caption{FFA-based simulated annealing \ffaSA}
            \label{fig:fsa-stn}
        \end{subfigure}
        \caption{Search trajectory networks for the QAP instance~\textbf{tai27e09}. For the purpose of visual clarity, only five runs are shown per plot. Each run is a different colour. Fitness is on the $y$-axis and solutions are placed in an ordering along the horizontal axis which is obtained from optimising their placement according to the relative permutation swap distances}
            \label{fig:stns}
    \end{figure}%

 \begin{figure}[h]
        \centering
        \begin{subfigure}[b]{0.49\textwidth}
            \centering
            \includegraphics[width=\textwidth]{supplemental/tai27e10-OHC.pdf}
            \caption{Objective-steered hill-climbing \objHC}
            \label{fig:ohc-stn}
        \end{subfigure}
        \hfill
        \begin{subfigure}[b]{0.49\textwidth}
            \centering
            \includegraphics[width=\textwidth]{supplemental/tai27e10-FHC.pdf}
            \caption{FFA-based hill-climbing \ffaHC}
            \label{fig:fhc-stn}
        \end{subfigure}
        \vskip\baselineskip
        \begin{subfigure}[b]{0.49\textwidth}
            \centering
            \includegraphics[width=\textwidth]{supplemental/tai27e10-OSA.pdf}
            \caption{Objective-steered SA \objSA}
            \label{fig:osa-stn}
        \end{subfigure}
        \hfill
        \begin{subfigure}[b]{0.49\textwidth}
            \centering
            \includegraphics[width=\textwidth]{supplemental/tai27e10-FSA.pdf}
            \caption{FFA-based simulated annealing \ffaSA}
            \label{fig:fsa-stn}
        \end{subfigure}
        \caption{Search trajectory networks for the QAP instance~\textbf{tai27e10}. For the purpose of visual clarity, only five runs are shown per plot. Each run is a different colour. Fitness is on the $y$-axis and solutions are placed in an ordering along the horizontal axis which is obtained from optimising their placement according to the relative permutation swap distances}
            \label{fig:stns}
    \end{figure}

     \begin{figure}[h]
        \centering
        \begin{subfigure}[b]{0.49\textwidth}
            \centering
            \includegraphics[width=\textwidth]{supplemental/tai27e11-OHC.pdf}
            \caption{Objective-steered hill-climbing \objHC}
            \label{fig:ohc-stn}
        \end{subfigure}
        \hfill
        \begin{subfigure}[b]{0.49\textwidth}
            \centering
            \includegraphics[width=\textwidth]{supplemental/tai27e11-FHC.pdf}
            \caption{FFA-based hill-climbing \ffaHC}
            \label{fig:fhc-stn}
        \end{subfigure}
        \vskip\baselineskip
        \begin{subfigure}[b]{0.49\textwidth}
            \centering
            \includegraphics[width=\textwidth]{supplemental/tai27e11-OSA.pdf}
            \caption{Objective-steered SA \objSA}
            \label{fig:osa-stn}
        \end{subfigure}
        \hfill
        \begin{subfigure}[b]{0.49\textwidth}
            \centering
            \includegraphics[width=\textwidth]{supplemental/tai27e11-FSA.pdf}
            \caption{FFA-based simulated annealing \ffaSA}
            \label{fig:fsa-stn}
        \end{subfigure}
        \caption{Search trajectory networks for the QAP instance~\textbf{tai27e11}. For the purpose of visual clarity, only five runs are shown per plot. Each run is a different colour. Fitness is on the $y$-axis and solutions are placed in an ordering along the horizontal axis which is obtained from optimising their placement according to the relative permutation swap distances}
            \label{fig:stns}
    \end{figure}%

 \begin{figure}[h]
        \centering
        \begin{subfigure}[b]{0.49\textwidth}
            \centering
            \includegraphics[width=\textwidth]{supplemental/tai27e12-OHC.pdf}
            \caption{Objective-steered hill-climbing \objHC}
            \label{fig:ohc-stn}
        \end{subfigure}
        \hfill
        \begin{subfigure}[b]{0.49\textwidth}
            \centering
            \includegraphics[width=\textwidth]{supplemental/tai27e12-FHC.pdf}
            \caption{FFA-based hill-climbing \ffaHC}
            \label{fig:fhc-stn}
        \end{subfigure}
        \vskip\baselineskip
        \begin{subfigure}[b]{0.49\textwidth}
            \centering
            \includegraphics[width=\textwidth]{supplemental/tai27e12-OSA.pdf}
            \caption{Objective-steered SA \objSA}
            \label{fig:osa-stn}
        \end{subfigure}
        \hfill
        \begin{subfigure}[b]{0.49\textwidth}
            \centering
            \includegraphics[width=\textwidth]{supplemental/tai27e12-FSA.pdf}
            \caption{FFA-based simulated annealing \ffaSA}
            \label{fig:fsa-stn}
        \end{subfigure}
        \caption{Search trajectory networks for the QAP instance~\textbf{tai27e12}. For the purpose of visual clarity, only five runs are shown per plot. Each run is a different colour. Fitness is on the $y$-axis and solutions are placed in an ordering along the horizontal axis which is obtained from optimising their placement according to the relative permutation swap distances}
            \label{fig:stns}
    \end{figure}%

 \begin{figure}[h]
        \centering
        \begin{subfigure}[b]{0.49\textwidth}
            \centering
            \includegraphics[width=\textwidth]{supplemental/tai27e13-OHC.pdf}
            \caption{Objective-steered hill-climbing \objHC}
            \label{fig:ohc-stn}
        \end{subfigure}
        \hfill
        \begin{subfigure}[b]{0.49\textwidth}
            \centering
            \includegraphics[width=\textwidth]{supplemental/tai27e13-FHC.pdf}
            \caption{FFA-based hill-climbing \ffaHC}
            \label{fig:fhc-stn}
        \end{subfigure}
        \vskip\baselineskip
        \begin{subfigure}[b]{0.49\textwidth}
            \centering
            \includegraphics[width=\textwidth]{supplemental/tai27e13-OSA.pdf}
            \caption{Objective-steered SA \objSA}
            \label{fig:osa-stn}
        \end{subfigure}
        \hfill
        \begin{subfigure}[b]{0.49\textwidth}
            \centering
            \includegraphics[width=\textwidth]{supplemental/tai27e13-FSA.pdf}
            \caption{FFA-based simulated annealing \ffaSA}
            \label{fig:fsa-stn}
        \end{subfigure}
        \caption{Search trajectory networks for the QAP instance~\textbf{tai27e13}. For the purpose of visual clarity, only five runs are shown per plot. Each run is a different colour. Fitness is on the $y$-axis and solutions are placed in an ordering along the horizontal axis which is obtained from optimising their placement according to the relative permutation swap distances}
            \label{fig:stns}
    \end{figure}%

 \begin{figure}[h]
        \centering
        \begin{subfigure}[b]{0.49\textwidth}
            \centering
            \includegraphics[width=\textwidth]{supplemental/tai27e14-OHC.pdf}
            \caption{Objective-steered hill-climbing \objHC}
            \label{fig:ohc-stn}
        \end{subfigure}
        \hfill
        \begin{subfigure}[b]{0.49\textwidth}
            \centering
            \includegraphics[width=\textwidth]{supplemental/tai27e14-FHC.pdf}
            \caption{FFA-based hill-climbing \ffaHC}
            \label{fig:fhc-stn}
        \end{subfigure}
        \vskip\baselineskip
        \begin{subfigure}[b]{0.49\textwidth}
            \centering
            \includegraphics[width=\textwidth]{supplemental/tai27e14-OSA.pdf}
            \caption{Objective-steered SA \objSA}
            \label{fig:osa-stn}
        \end{subfigure}
        \hfill
        \begin{subfigure}[b]{0.49\textwidth}
            \centering
            \includegraphics[width=\textwidth]{supplemental/tai27e14-FSA.pdf}
            \caption{FFA-based simulated annealing \ffaSA}
            \label{fig:fsa-stn}
        \end{subfigure}
        \caption{Search trajectory networks for the QAP instance~\textbf{tai27e14}. For the purpose of visual clarity, only five runs are shown per plot. Each run is a different colour. Fitness is on the $y$-axis and solutions are placed in an ordering along the horizontal axis which is obtained from optimising their placement according to the relative permutation swap distances}
            \label{fig:stns}
    \end{figure}%

 \begin{figure}[h]
        \centering
        \begin{subfigure}[b]{0.49\textwidth}
            \centering
            \includegraphics[width=\textwidth]{supplemental/tai27e15-OHC.pdf}
            \caption{Objective-steered hill-climbing \objHC}
            \label{fig:ohc-stn}
        \end{subfigure}
        \hfill
        \begin{subfigure}[b]{0.49\textwidth}
            \centering
            \includegraphics[width=\textwidth]{supplemental/tai27e15-FHC.pdf}
            \caption{FFA-based hill-climbing \ffaHC}
            \label{fig:fhc-stn}
        \end{subfigure}
        \vskip\baselineskip
        \begin{subfigure}[b]{0.49\textwidth}
            \centering
            \includegraphics[width=\textwidth]{supplemental/tai27e15-OSA.pdf}
            \caption{Objective-steered SA \objSA}
            \label{fig:osa-stn}
        \end{subfigure}
        \hfill
        \begin{subfigure}[b]{0.49\textwidth}
            \centering
            \includegraphics[width=\textwidth]{supplemental/tai27e15-FSA.pdf}
            \caption{FFA-based simulated annealing \ffaSA}
            \label{fig:fsa-stn}
        \end{subfigure}
        \caption{Search trajectory networks for the QAP instance~\textbf{tai27e15}. For the purpose of visual clarity, only five runs are shown per plot. Each run is a different colour. Fitness is on the $y$-axis and solutions are placed in an ordering along the horizontal axis which is obtained from optimising their placement according to the relative permutation swap distances}
            \label{fig:stns}
    \end{figure}%

 \begin{figure}[h]
        \centering
        \begin{subfigure}[b]{0.49\textwidth}
            \centering
            \includegraphics[width=\textwidth]{supplemental/tai27e16-OHC.pdf}
            \caption{Objective-steered hill-climbing \objHC}
            \label{fig:ohc-stn}
        \end{subfigure}
        \hfill
        \begin{subfigure}[b]{0.49\textwidth}
            \centering
            \includegraphics[width=\textwidth]{supplemental/tai27e16-FHC.pdf}
            \caption{FFA-based hill-climbing \ffaHC}
            \label{fig:fhc-stn}
        \end{subfigure}
        \vskip\baselineskip
        \begin{subfigure}[b]{0.49\textwidth}
            \centering
            \includegraphics[width=\textwidth]{supplemental/tai27e16-OSA.pdf}
            \caption{Objective-steered SA \objSA}
            \label{fig:osa-stn}
        \end{subfigure}
        \hfill
        \begin{subfigure}[b]{0.49\textwidth}
            \centering
            \includegraphics[width=\textwidth]{supplemental/tai27e16-FSA.pdf}
            \caption{FFA-based simulated annealing \ffaSA}
            \label{fig:fsa-stn}
        \end{subfigure}
        \caption{Search trajectory networks for the QAP instance~\textbf{tai27e16}. For the purpose of visual clarity, only five runs are shown per plot. Each run is a different colour. Fitness is on the $y$-axis and solutions are placed in an ordering along the horizontal axis which is obtained from optimising their placement according to the relative permutation swap distances}
            \label{fig:stns}
    \end{figure}%

 \begin{figure}[h]
        \centering
        \begin{subfigure}[b]{0.49\textwidth}
            \centering
            \includegraphics[width=\textwidth]{supplemental/tai27e17-OHC.pdf}
            \caption{Objective-steered hill-climbing \objHC}
            \label{fig:ohc-stn}
        \end{subfigure}
        \hfill
        \begin{subfigure}[b]{0.49\textwidth}
            \centering
            \includegraphics[width=\textwidth]{supplemental/tai27e17-FHC.pdf}
            \caption{FFA-based hill-climbing \ffaHC}
            \label{fig:fhc-stn}
        \end{subfigure}
        \vskip\baselineskip
        \begin{subfigure}[b]{0.49\textwidth}
            \centering
            \includegraphics[width=\textwidth]{supplemental/tai27e17-OSA.pdf}
            \caption{Objective-steered SA \objSA}
            \label{fig:osa-stn}
        \end{subfigure}
        \hfill
        \begin{subfigure}[b]{0.49\textwidth}
            \centering
            \includegraphics[width=\textwidth]{supplemental/tai27e17-FSA.pdf}
            \caption{FFA-based simulated annealing \ffaSA}
            \label{fig:fsa-stn}
        \end{subfigure}
        \caption{Search trajectory networks for the QAP instance~\textbf{tai27e17}. For the purpose of visual clarity, only five runs are shown per plot. Each run is a different colour. Fitness is on the $y$-axis and solutions are placed in an ordering along the horizontal axis which is obtained from optimising their placement according to the relative permutation swap distances}
            \label{fig:stns}
    \end{figure}%

 \begin{figure}[h]
        \centering
        \begin{subfigure}[b]{0.49\textwidth}
            \centering
            \includegraphics[width=\textwidth]{supplemental/tai27e18-OHC.pdf}
            \caption{Objective-steered hill-climbing \objHC}
            \label{fig:ohc-stn}
        \end{subfigure}
        \hfill
        \begin{subfigure}[b]{0.49\textwidth}
            \centering
            \includegraphics[width=\textwidth]{supplemental/tai27e18-FHC.pdf}
            \caption{FFA-based hill-climbing \ffaHC}
            \label{fig:fhc-stn}
        \end{subfigure}
        \vskip\baselineskip
        \begin{subfigure}[b]{0.49\textwidth}
            \centering
            \includegraphics[width=\textwidth]{supplemental/tai27e18-OSA.pdf}
            \caption{Objective-steered SA \objSA}
            \label{fig:osa-stn}
        \end{subfigure}
        \hfill
        \begin{subfigure}[b]{0.49\textwidth}
            \centering
            \includegraphics[width=\textwidth]{supplemental/tai27e18-FSA.pdf}
            \caption{FFA-based simulated annealing \ffaSA}
            \label{fig:fsa-stn}
        \end{subfigure}
        \caption{Search trajectory networks for the QAP instance~\textbf{tai27e18}. For the purpose of visual clarity, only five runs are shown per plot. Each run is a different colour. Fitness is on the $y$-axis and solutions are placed in an ordering along the horizontal axis which is obtained from optimising their placement according to the relative permutation swap distances}
            \label{fig:stns}
    \end{figure}%

 \begin{figure}[h]
        \centering
        \begin{subfigure}[b]{0.49\textwidth}
            \centering
            \includegraphics[width=\textwidth]{supplemental/tai27e19-OHC.pdf}
            \caption{Objective-steered hill-climbing \objHC}
            \label{fig:ohc-stn}
        \end{subfigure}
        \hfill
        \begin{subfigure}[b]{0.49\textwidth}
            \centering
            \includegraphics[width=\textwidth]{supplemental/tai27e19-FHC.pdf}
            \caption{FFA-based hill-climbing \ffaHC}
            \label{fig:fhc-stn}
        \end{subfigure}
        \vskip\baselineskip
        \begin{subfigure}[b]{0.49\textwidth}
            \centering
            \includegraphics[width=\textwidth]{supplemental/tai27e19-OSA.pdf}
            \caption{Objective-steered SA \objSA}
            \label{fig:osa-stn}
        \end{subfigure}
        \hfill
        \begin{subfigure}[b]{0.49\textwidth}
            \centering
            \includegraphics[width=\textwidth]{supplemental/tai27e19-FSA.pdf}
            \caption{FFA-based simulated annealing \ffaSA}
            \label{fig:fsa-stn}
        \end{subfigure}
        \caption{Search trajectory networks for the QAP instance~\textbf{tai27e19}. For the purpose of visual clarity, only five runs are shown per plot. Each run is a different colour. Fitness is on the $y$-axis and solutions are placed in an ordering along the horizontal axis which is obtained from optimising their placement according to the relative permutation swap distances}
            \label{fig:stns}
    \end{figure}%

 \begin{figure}[h]
        \centering
        \begin{subfigure}[b]{0.49\textwidth}
            \centering
            \includegraphics[width=\textwidth]{supplemental/tai27e20-OHC.pdf}
            \caption{Objective-steered hill-climbing \objHC}
            \label{fig:ohc-stn}
        \end{subfigure}
        \hfill
        \begin{subfigure}[b]{0.49\textwidth}
            \centering
            \includegraphics[width=\textwidth]{supplemental/tai27e20-FHC.pdf}
            \caption{FFA-based hill-climbing \ffaHC}
            \label{fig:fhc-stn}
        \end{subfigure}
        \vskip\baselineskip
        \begin{subfigure}[b]{0.49\textwidth}
            \centering
            \includegraphics[width=\textwidth]{supplemental/tai27e20-OSA.pdf}
            \caption{Objective-steered SA \objSA}
            \label{fig:osa-stn}
        \end{subfigure}
        \hfill
        \begin{subfigure}[b]{0.49\textwidth}
            \centering
            \includegraphics[width=\textwidth]{supplemental/tai27e20-FSA.pdf}
            \caption{FFA-based simulated annealing \ffaSA}
            \label{fig:fsa-stn}
        \end{subfigure}
        \caption{Search trajectory networks for the QAP instance~\textbf{tai27e20}. For the purpose of visual clarity, only five runs are shown per plot. Each run is a different colour. Fitness is on the $y$-axis and solutions are placed in an ordering along the horizontal axis which is obtained from optimising their placement according to the relative permutation swap distances}
            \label{fig:stns}
    \end{figure}%

%